\documentclass{article}
\usepackage[utf8]{inputenc}

\usepackage[english]{babel}
\usepackage[utf8]{inputenc}
\usepackage[T1]{fontenc}

\usepackage{enumitem}

\usepackage[a4paper,top=3cm,bottom=2cm,left=3cm,right=3cm,marginparwidth=1.75cm]{geometry}

\usepackage{amsmath}
\usepackage{verbatim}
\usepackage{amssymb}
\usepackage{mathrsfs}
\usepackage{stmaryrd}
\usepackage{graphicx}
\usepackage[singlelinecheck=false]{caption} 
\usepackage[square,comma,numbers,sort&compress]{natbib}
\usepackage{bbold}
\usepackage[colorinlistoftodos]{todonotes}
\usepackage[colorlinks=true, allcolors=blue]{hyperref}
\usepackage[french,onelanguage]{algorithm2e}
\usepackage{multirow,tabularx}
\usepackage{diagbox}
\usepackage{lscape}
\hypersetup{colorlinks,%
citecolor=blue,%
filecolor=black,%
linkcolor=black,%
urlcolor=black
}
\usepackage[super]{nth}
\usepackage{pdfpages}
\newcommand{\distas}[1]{\mathbin{\overset{#1}{\kern\z@\sim}}}%

\DeclareMathOperator*{\argmin}{arg\!\min}

\begin{document}

\title{\textbf{Impact of Low-bitwidth Quantization on the Adversarial Robustness for Embedded Neural Networks}}
\author{Rémi Bernhard\textsuperscript{1}, Pierre-Alain Moellic\textsuperscript{1}, Jean-Max Dutertre\textsuperscript{2} \\ \\
\textsuperscript{1}CEA Tech, Systemes et Architectures S\'ecuris\'ees (SAS),\\Centre CMP, Equipe Commune CEA Tech - Mines Saint-Etienne\\ Gardanne, France \\ 
\textsuperscript{2}Mines Saint-Etienne, CEA-Tech,\\ Centre CMP,\\
Gardanne, France \\ \\
\textbf{remi.bernhard@cea.fr, pierre-alain.moellic@cea.fr, dutertre@emse.fr}
}



\date{}
\maketitle

\section{Abstract}
\label{Abstract}
As the will to deploy neural networks models on embedded systems grows, and considering the related memory footprint and energy consumption issues, finding lighter solutions to store neural networks such as weight quantization and more efficient inference methods become major research topics. 
Parallel to that, adversarial machine learning has risen recently with an impressive and significant attention, unveiling some critical flaws of machine learning models, especially neural networks. In particular, perturbed inputs called \textit{adversarial examples} have been shown to fool a model into making incorrect predictions.\\
In this article, we investigate the adversarial robustness of quantized neural networks under different threat models for a classical supervised image classification task. We show that quantization does not offer any robust protection, results in severe form of gradient masking and advance some hypotheses to explain it. However, we experimentally observe poor transferability capacities which we explain by \textit{quantization value shift} phenomenon and gradient misalignment and explore how these results can be exploited with an ensemble-based defense.

\section{Introduction}
\label{Introduction}
\subsection{Context}
\label{Context}
    
Neural networks achieve state-of-the art performances in various domains such as speech translation or image recognition. These outstanding performances have been allowed~--~among others~--~by tremendous computation power (e.g., popularization of GPU) and the resulting trained architectures come with thousands or even millions parameters. 
As the desire to run pre-trained neural network based application (e.g image recognition) on embedded or mobile systems grows, one must investigate the ways to solve the practical issues involved. First of all, the memory footprint can quickly be a limiting factor for constrained devices. For example, a typical ARM Cortex-M4-based microcontroller such as STM32F4 has up to 384 KBytes RAM and a maximum of 2MBytes of Flash memory\footnote{https://www.st.com/en/microcontrollers-microprocessors/stm32f4-series.html}. Secondly, inference cost in terms of energy is critical for devices like mobile phones or a large variety of connected objects (e.g., industrial sensors). Thirdly, inference speed is necessary to avoid critical latency issues. 

Some APIs like the Android Neural Network API (NNAPI\footnote{https://developer.android.com/ndk/guides/neuralnetworks}) have been already developed to allow to run efficiently trained models with famous frameworks (TensorFlow\footnote{https://www.tensorflow.org/}, etc.) on Android systems. Tensorflow Lite (TFLite\footnote{https://www.tensorflow.org/lite}) allows to transfer pre-trained model to mobile or embedded devices thanks to model compression techniques and 8-bit post-training weights quantization. ARM-NN\footnote{https://developer.arm.com/products/processors/machine-learning/arm-nn} is another SDK that does the link for applications between various machine learning frameworks and diverse Cortex CPUs or Mali GPUs types (note that CMSIS-NN\footnote{https://github.com/ARM-software/CMSIS\_5} is dedicated to Cortex-M MCU). STMicroelectronics also proposes an AI expansion pack for the STM32CubeMX, called X.Cube.AI\footnote{https://www.st.com/en/embedded-software/x-cube-ai.html}, to map pre-trained neural network models into different STM32 microcontroller series thanks to 8-bit post-training quantization and other optimization tricks related to the specificity of these platforms.

On a more theoretical side, research about reducing the number of parameters to directly impact the memory footprint of models \citep{denil2013predicting,hacene2018quantized, gong2014compressing,han2015deep,choi2016towards}, or developing quantization schemes coupled with efficient computation methods to reduce inference time and energy consumption has arisen \citep{courbariaux2015binaryconnect,courbariaux2016binarized,hubara2017quantized,rastegari2016xnor,li2016ternary,zhu2016trained,gupta2015deep,zhou2016dorefa,ding2017lightnn,polino2018model}.\\
At the same time, neural networks have been shown to be vulnerable to malicious tampering of inputs \citep{szegedy2013intriguing}. From a clean observation correctly classified by a model, an adversary optimally crafts a so-called \textit{adversarial example}, which is very similar to the clean observation and fools the model. Many attack methods (\citep{goodfellow2015laceyella,carlini2017towards,moosavi2016deepfool,kurakin2016adversarial,papernot2017practical,chen2017zoo} for some of the most famous) and defense methods (\citep{Madry2017,szegedy2013intriguing,dhillon2018stochastic,metzen2017detecting,grosse2017statistical} for some of the most famous) have been developed and evaluated in benchmarks or competition tracks such as the NIPS \textit{Adversarial Vision Challenge}\footnote{https://www.crowdai.org/challenges/adversarial-vision-challenge}.

\subsection{Motivation and related works}
\label{Motivation}

In terms of security, as embedded systems with neural networks models become ubiquitous, it is a particularly interesting topic to evaluate the robustness of state-of-the-art quantization methods under different threat models. Moreover, studying the transferability of adversarial examples between original (i.e. full precision) and quantized neural networks may at the same time highlight weaknesses or strengths of future embedded systems, and allow to better understand if quantization in itself could be a relevant defense against adversarial examples or, on the contrary, exacerbates these flaws.

Some authors have already investigated the link between quantization and robustness. \citep{galloway2017attacking} claimed that neural networks trained with weights and activation values binarized to $\{-1,1\}$ have an interesting robustness against adversarial examples. However, this robustness was demonstrated thanks to the only MNIST dataset 
and use stochastic quantization. This quantization scheme induces the stochastic gradient phenomenon \citep{athalye2018obfuscated}, which can mislead to the true efficiency of this defense by causing what \citep{uesato2018adversarial} called \textit{obscurity}. \citep{lin2018defensive} tries to explain some weaknesses of quantization-based defense methods against adversarial examples. They show experimentally that these defense methods can, in fact, denoise an adversarial example or enlarge its perturbation, depending on the size of the perturbation in the input space and the number of bits used for quantization. Thus, quantization can participate in an error amplification or attenuation effect. However, they only apply the FGSM attack~\citep{goodfellow2015laceyella} in a white-box setting against simple activation quantization.
Although focused on model compression (pruning),~\citep{zhao2018compress} studied the robustness of quantized neural networks against adversarial examples with a fixed-point quantization scheme applies to both weight and activation values, no less than 4-bit model, and a restricted set of (gradient-based) attacks.
\citep{rakin2018defend} proposes a defense method based on activation quantization coupled with adversarial training \citep{Madry2017}, which has been shown by \citep{lin2018defensive} to introduce gradient masking \citep{Papernot2016}.
Interestingly, \citep{khalil2018combinatorial} notes that the gradients obtained via the use of a \textit{Straight Through Estimator} (hereafter STE, \citep{bengio2013estimating})~--~a common technique to compute gradients when quantization operations lead to differentiability issues~--~may not be representative of the true gradient. This observation leads to questions about efficiency of gradient-based attacks against quantized neural networks, and strengthens up the motivation to study gradient masking issues and black-box attacks against such models. The authors propose a Mixed Integer Linear Programming (MILP) based attack, which shows good results on the MNIST data set but is not scalable to large neural networks, due to computation cost issues. \\

\subsection{Contributions}
\label{Contributions}

In this work, we study the robustness of natural and quantized models and against adversarial examples under different threat models against various types of attacks. Our contributions are:
\begin{itemize}
    \item[--] We show that quantization in itself offers poor protection against various well-known adversarial crafting methods and we explain why activation quantization can lead to severe gradient masking, a phenomenon which leads to non-useful gradients  to craft adversarial examples \citep{papernot2017practical} and causes ineffective defense \citep{uesato2018adversarial}.
    \item[--] We show very poor transferability capacities of adversarial examples between full-precision and quantized models and between quantized models with different bitwidths. We advance hypothesis to explain it, including a quantization shift phenomenon and gradient misalignment.
    \item[--] We investigate a defense based on an ensemble of models with different quantization levels. Results are promising and pave the way towards a possible perspective for future work.
\end{itemize}

\section{Background}
\label{Background}
\subsection{Quantization of neural networks}
\label{Quantization of neural networks}

The purpose of this article is to study the impact of quantization techniques on the adversarial robustness, for embedded neural networks. However, other complementary approaches are extensively studied to compress as well as to speed up models at inference time. During inference, energy consumption grows with memory access, which itself grows with memory footprint. Thus, reducing the number of parameters has been logically investigated. For example, \citep{denil2013predicting} show, for specific architectures and datasets, that some of the parameters are predictable from the others.
 \citep{han2015deep} develop a three-step method (pruning, clustering, tuning) to efficiently compress a neural network achieving a reduction of AlexNet memory footprint by a factor of 35. \citep{hacene2018quantized} propose a method to reduce the memory size of a convolutional neural network by pruning connections based on a deterministic rule. This method is also coupled with weight binarization (\citep{courbariaux2015binaryconnect}) and an efficient hardware architecture on a FPGA in order to reduce inference time.\\



Reducing the precision of the weights or developing efficient computation methods for some precise format of weight values is an important field of investigation. Quantization can be performed as a post-training process or during training. For the first case, as previously described in introduction, several tools have been recently proposed to map full precision pre-trained models for inference purpose (TFLite, ARM-NN, STMCubeMX A.I.\footnote{https://www.st.com/en/embedded-software/x-cube-ai.html}) by coarsely quantizing some weights into~--~usually~--~no more than 8-bit integers. More advanced methods propose clustering methods~\citep{choi2016towards} or information theoretical vector quantization methods (inspired by \citep{denil2013predicting}) such as \citep{gong2014compressing} who achieved 
 about 20 times compression of the model with only 1\% loss of
classification accuracy on the Imagenet benchmark. 

In this article, we focus our work on quantization techniques at training time since these approaches enable to reach state-of-the-art performance with lower bitwidth precision. Hereunder, we detail some of the most popular works on that field that we consider for our experimentations.\\

\textbf{Binary Connect and Binary Net.}
\label{Binary Connect and Binary Net}
\citep{courbariaux2015binaryconnect} presents a method to train neural networks with weights $w$ binarized to $w_b \in \{-1,1\}$. During training, weights are binarized for the forward pass, and as the binarization operation can be not differentiable or lead to the vanishing gradient problem, the STE given in Equation \ref{eq_ste} is used for the backward pass:
\begin{equation}
    \frac {\partial C(w)} {\partial w} \approx  \frac {\partial C} {\partial w} \bigg\rvert_{\substack{w=w_b}} \textbf{1}_{|w|\leq1} 
    \label{eq_ste}
\end{equation}
Where $C$ is the cost function and \textbf{1}(.) the indicator function. \citep{courbariaux2016binarized} pursue this idea by training binary networks (BNN) with weight values $w$ and activation function $a^k$ values binarized to $ (w_b, a_b^k) \in \{-1,1\}^2$. During the backward pass, the authors used the same STE principle for activations as  in Equation \ref{eq_ste} above. Improvements have been proposed as in \citep{darabi2018bnn} by adding regularization and more complex approximation of the derivative on the backward pass.
\\

\textbf{Xnor Net.}
\citep{rastegari2016xnor} binarizes weigths and activation values no more to $\{-1,1\}$ but to $\{-\alpha , \alpha\}$ with $\alpha  \in \mathbb{R}^{*,+}$. They formalize the search of the best binarization approximation of the real-valued weights as the following optimization problem.:
\begin{equation}
    \argmin_{B, \alpha} \left\| W - \alpha B \right\|_2
\end{equation}
Where $W$ is the weight matrix and $B$ is a matrix with only -1 and 1.
During the backward pass, a STE is used.\\

\textbf{Ternarization.}
In \citep{li2016ternary}, the authors propose a method to train a neural network with weight values ternarized to $\{-1,0,1\}$ during the forward pass.~\citep{zhu2016trained} also propose a method to train a neural network with weight values ternarized to $\{-\alpha_l,0,\alpha_u\}$ during the forward pass, where $\alpha_l$, $\alpha_u$ $\in \mathbb{R}^{2,*}_{+}$ and $\alpha_l$ and $\alpha_u$ are updated during training.\\

\textbf{Low bitwidth quantization.}
\label{Low bitwidth quantization}
\citep{gupta2015deep} successfully train networks with good precisions on MNIST and CIFAR10 data sets while limiting the bitwidth of the weights values to 16 bits and using stochastic rounding.~\citep{zhou2016dorefa} proposes a method to train neural networks with low-bitwidth weight values, gradients and activation function values.   
They claim that taking advantage of this technique during the forward pass could help speed up the training of neural network on resource-limited hardware, and naturally speed up the inference. 
For a real value $x \in \left[0,1 \right]$ and a number of bits $n$, the function 
\begin{equation}
    Q(x,n) = \frac{round((2^n -1) x)}{2^n -1}
\end{equation}
is the quantization function used for weights, activation values and gradients. The weight and activation values are quantized on the forward pass only. The authors also found that quantizing gradients on the backward pass (with a STE) was requisite. $Q(x,n)  (2^n -1)$ results in $2^n$ values, each of them representable with a $n$ bits integer. During the forward pass, one can take advantage of the bit convolution kernel method (see \citep{zhou2016dorefa} for details) with respect to the $Q(x,n)  (2^n -1)$ values, and then scale afterwards with the $2^n -1$ value.\\
\citep{ding2017lightnn} proposes another weight quantization method with a constraint on the number of "1" in the binary representation of weights, along with some efficient computation method.\\
\citep{polino2018model} proposes a method which involves distillation and quantization of the weight values to decrease the storage size of a model. For the quantization part, given some weight value $w$, it is first scaled to a value $v$ in $\left[ 0,1 \right]$, then mapped with a function $Q$ to the nearest mapping points among the $s+1$ points in $\left[0,1\right]$ and then scaled back to the original scale. \\

In this article, we use the Binary Net method and Binary Connect respectively from \citep{courbariaux2016binarized} and \citep{courbariaux2015binaryconnect}, and the quantization method (Dorefa-Net) from \citep{zhou2016dorefa}.\\

    \subsection{Adversarial machine learning}
    \label{Adversarial machine learning}
    
Machine learning systems have been shown to be vulnerable against different types of attacks threatening their confidentiality, integrity or accessibility. We can distinguish three different types of attacks: 
\begin{itemize}[label=--]
	\item \textit{Data/Model leakage} occurs once the model has been trained. An adversary aims at stealing models parameters or architecture, or stealing confidential or private (training) data \citep{fredrikson2015model, shokri2017membership,shokri2015privacy}.
	\item With \textit{data poisoning}, which steps in during the training phase, an attacker targets the integrity or availability of the system according to the level of the perturbation. In order to decrease the model accuracy, corrupted data are introduced in the training set when the data are collected in the physical world or directly in the model input domain \citep{munoz2017towards,yang2017generative}.
	\item Adversaries may also alter the inputs at inference time, by crafting malicious observations looking like clean ones but designed such as to fool the model \citep{szegedy2013intriguing,goodfellow2015laceyella}, striking the model integrity.
\end{itemize}
Here we focus on the latter type of integrity-based attack, i.e \textit{adversarial examples crafting}.\\

\subsubsection{Adversarial examples}
\label{Adversarial examples}     

Adversarial examples are highly worrying threats to machine learning. Roughly stated, considering a classifier model, an adversarial example is a slightly modified version of a correctly classified clean example in a way such that the classifier will output two different classes for those two examples.\\
The reason of existence of adversarial examples lead to various hypothesis.
\citep{szegedy2013intriguing} propose a first explanation to the existence of adversarial examples: they would be in fact located in low-probability pockets of the input space. On their side, for \citep{goodfellow2015laceyella}, it is some local linearity assumption which eases the crafting of adversarial examples, and not the global non-linear nature of neural networks. In \citep{tanay2016boundary}, the authors give a more geometric explanation to the existence of adversarial examples, saying that the learned boundary "extends beyond the submanifold of sample data and can be~--~under certain circumstances~--~lying close to it" (\textit{boundary tilting} effect), and argue that the linearity assumption is not sufficient to explain adversarial examples that are~--~for some of them~--~the result of overfitting issues. \citep{gilmer2018adversarial}, based on theoretical results linking the generalization error to the average distance to a misclassified point for a very particular type of dataset, include high dimensionality as a possible power factor of adversarial examples. Another common hypothesis, used among others to detect adversarial examples, is that adversarial examples are not on the data manifold \citep{samangouei2018defense}. Recently, \citep{ilyas2019adversarial} show that adversarial examples are the consequence of \textit{non-robust features} derived from patterns with a big predictive power, yet these patterns are meaningless to humans and they can be adversarially modified to fool the target classifier.\\
More precisely, given a classifier model $M$ learning a mapping function $f: \mathbb{R}^m \xrightarrow{} \{1,...,K \}$, given an initial clean observation $x \in \mathbb{R}^m$, given a target label $t \in \{1,...,K \}$,  a targeted adversarial example $x ' \in \mathbb{R}^m$ crafted from a correctly classified $x$ is defined such as $f(x) \neq f(x')=t$ and $d(x,x') < \epsilon$ with $d$ a distance function being often the distance derived from the $l_2$ or $l_{\infty}$ norm.\\
Based on \citep{szegedy2013intriguing}, the search of such an adversarial example can be written as:
\\
\begin{align*}
    \epsilon &  = \argmin_{\epsilon}{ \left\| \epsilon \right\|_p} \\
    s.t \quad f(x + \epsilon) & = t \text{~(\textit{targeted attack)}} \\
    s.t \quad f(x + \epsilon) & \neq f(x)  \text{~(\textit{untargeted attack)}}
\end{align*}

Usually, the adversary may also wants that $x + \epsilon$ be bounded (for example, $x+ \epsilon \in [0,1]$ as in \citep{szegedy2013intriguing}).\\


\subsubsection{Attacks}
\label{Attacks}

In this article, we use five different adversarial crafting methods. These attacks are presented in their untargeted version, where adversarial examples are crafted from a clean observation $x$ of label $y$, $logit_j(x)$ designates the logit output for the $j^{th}$ class, and $f_j(x)$ designates the softmax output for the $j^{th}$ class.\\

\textbf{Fast Gradient Sign Method (FGSM).}
Presented by \citep{goodfellow2015laceyella}, this method, derives an adversarial example $x$ maximizing $J(\theta, x + \alpha,y) - J(\theta,x,y)$ with respect to $\alpha$ ,given that $\left\| \alpha \right\|_{\infty} < \epsilon$,  by performing a linear approximation of the loss function $J(\theta,x,y)$ around $x$, one gets:
\begin{equation}
x' = x + \epsilon Sign( \frac{\partial J}{\partial x}(\theta,x,y))
\end{equation}
$x'$ is then clipped to respect a possible box constraint (for images for example one may want $x' \in [0,1]$)\\

\textbf{Basic Iterative Method (BIM)}~\citep{kurakin2016adversarial2} presents the Basic Iterative Method, derived from the FGSM method, which allows to craft targeted adversarial perturbations. Given a maximum adversarial perturbation $\epsilon$:
\begin{equation}
     x_0 = x,~
     x_{n+1} = Clip_{B_{\infty}(x, \epsilon)}(x_n + \alpha sign( \frac{\partial J}{\partial x}(\theta,x_n,y)))
\end{equation}

where $B_{\infty}(x, \epsilon)$ is the $l_{\infty}$-ball of radius $\epsilon$ and center $x$, and we set $\alpha = \frac{\epsilon}{T}$ with $T$ the total number of iterations.
In fact, we just repeat the targeted FGSM method for $K$ iterations, performing clipping at each iteration.
$x'$ is then clipped to respect a possible box constraint (for images for example one must have $x' \in [0,1]$).\\

\textbf{Carlini-Wagner $l_2$ (CWl2).}
Presented by \citep{carlini2017towards}, the Carlini-Wagner $l_2$ method consists of considering the following objective:\\
\begin{align}
    \min_{\epsilon} \quad & \left\| \epsilon \right\|_2 + c F(x + \epsilon,y),\nonumber\\    
    & s.t. \quad x + \epsilon \in [0,1]
    \label{eq_cwl2}
\end{align}
where:
\begin{equation}
F(x + \epsilon, y) = max(logit_y(x + \epsilon) - max_{j \neq y} logit_j(x + \epsilon), -\kappa)  
\end{equation}
with $\kappa \geq 0, c >0$. We set $\kappa=0$ and thus we have $F(x+\epsilon,y) = 0 \iff  label(x) \neq y$.
$c \in \mathbb{R}^{+}$ is a constant for which binary search is performed a decided amount of time. Then, the change of variable $x=\frac{1}{2}(tanh(w)+1)$ is performed to get rid of the box constraint. The resulting optimization problem with respect to the new variable $w$ can then be solved with classical optimization methods like Stochastic Gradient Descent (SGD) or Adam.\\

\textbf{SPSA attack.}
In \citep{uesato2018adversarial}, the authors propose a very effective gradient-free attack to evaluate defense strategies. The authors propose the constrained optimization problem given in Equation \ref{eq_spsa} that they solve by using the Adam update rule,approximating the gradients with finite difference estimates thanks to the SPSA (\textit{Simultaneous Perturbation Stochastic Approximation}, \citep{spall1992multivariate}) technique which is suitable for noisy high dimensional optimization problems, and performing clipping at each iteration to respect the constraint $ \left\| \alpha  \right\|_{\infty} < \epsilon$: 
\begin{align}
    \min_{\epsilon} \quad & logit_{y}(x + \alpha) - max_{j \neq y} logit_j(x + \alpha),\nonumber\\
    & s.t. \left\| \alpha  \right\|_{\infty} < \epsilon
\label{eq_spsa}
\end{align}

\textbf{Zeroth Order Optimization (ZOO).}
The ZOO attack \citep{chen2017zoo} is based on the CWl2 attack with a discrete approximation of the gradients:
\begin{equation}
    g'(x)_i \simeq \frac{g(x + h e_i)   - g(x - h e_i) }{2 h}    
\end{equation}
where $e_i$ is the basis vector with only the $i^{th}$ element equal to 1, the others equal 0 and $h$ is a small constant. The ZOO attack does not consider the logits values as the CWl2 attack does but the logarithm of the softmax output values, i.e we have:
\begin{equation}
    F_{ZOO}(x + \epsilon, y) = max(log(f_y(x + \epsilon)) - max_{j \neq y} log(f_j(x + \epsilon)), -\kappa)  
\end{equation}

\vspace{0.5cm}

\textbf{Characteristics.}
We sum up the main characteristics of these attacks in table \ref{attacks characteristics}.

\begin{table}[h!]
\begin{center}
\begin{tabularx}{\linewidth}{|c|*{5}{>{\centering \arraybackslash}X|}}
\hline
& FGSM & BIM & CWL2 & SPSA & ZOO \\ 
\hline
\textbf{Gradient-based} & \checkmark & \checkmark & \checkmark & & \\
\hline
\textbf{Gradient-free} &  & & & \checkmark & \checkmark\\
\hline
\textbf{one-step} & \checkmark & & & & \\
\hline
\textbf{iterative} & & \checkmark & \checkmark & \checkmark & \checkmark\\
\hline
$\boldsymbol{l_{\infty}}$ & \checkmark & \checkmark & & \checkmark & \\
\hline
$\boldsymbol{l_2}$ & & &\checkmark & & \checkmark \\
\hline
\end{tabularx}
\end{center}
\caption{Main characteristics of the considered adversarial examples crafting methods}
\label{attacks characteristics}
\end{table}

\subsubsection{Defenses}
\label{Defenses}

Many defenses have been investigated to counter adversarial examples. As the core of this article does not either aim at testing defense schemes or a specific attack, we refer to \citep{serban2018adversarial} for an overview of protections. The authors distinguish mainly the reactive defenses, which encompass pre-processing inputs and detection methods \citep{buckman2018thermometer,xu2017feature,xie2017mitigating,grosse2017statistical,feinman2017detecting,zheng2018robust,gong2017adversarial,metzen2017detecting,samangouei2018defense,meng2017magnet,lu2017safetynet}, the proactive defenses, which encompass techniques to make a network in itself more robust to adversarial examples \citep{goodfellow2015laceyella,Madry2017,tramer2017ensemble,chang2018efficient,kannan2018adversarial,zheng2018pgdadversarial,zhang2019theoretically,dhillon2018stochastic,kariyappa2019improving}, and provable defense methods \citep{raghunathan2018certified,kolter2017provable,hein2017formal, peck2017lower,gowal2018effectiveness}.

\subsection{Threat model}
\label{Threat model}

\subsubsection{Main characteristics of threat models}

The threat model encompasses assumptions about the adversary's goals, capabilities and knowledge.

\paragraph{Adversarial goal}
\label{Adversarial goal}
            
Here we focus on an adversary that aims to fool a supervised model at inference time. From a clean observation $x$ correctly labeled as $y$, the adversary wants to craft an adversarial example $x'$ labeled as a precise class $t \neq y$ (\textit{targeted attacks}) or any class $y' \neq y$ (\textit{untargeted attacks}). Given some threat model, a defense method claiming robustness against untargeted attacks is stronger than a one claiming robustness against targeted attacks. Similarly, it is often more difficult to craft targeted adversarial examples than untargeted ones.

\paragraph{Adversarial capability}
\label{Adversarial Capability}

It is crucial to properly define how much an adversary can alter a process of the machine learning pipeline. In the scope of adversarial examples crafting, the adversary's ability is almost all the time defined as an upper bound $\epsilon$ of the distance $D(x, x')$ between a clean observation $x$ and the adversarial example $x'$ crafted from $x$. The distance $D$ is derived from an $l_p$ norm, usually the $l_0$, $l_1$, $l_2$ or $l_{\infty}$ norm.
        
\paragraph{Adversarial knowledge}
\label{Adversarial Knowledge}
        
Traditionally, two main different settings are used to describe the way an adversary can operate. Each setting contains its own nuances but for the sake of simplicity we only present those we will later consider in this article. 

In the \textit{white-box} setting, the adversary is assumed to have a full access to the target model. This includes the type of model (SVM, architecture of a neural network, etc.), the parameters of the model (network's weights, etc.), any preprocessing component, etc. 

In a rigorous \textit{black-box} setting, the adversary has no information about the model but can (only) query it (in a limited or unlimited way). However, this setting can be loosened (some talk about \textit{grey-box} settings) according to the kind of information the adversary can get when querying the model (full prediction outputs~--~softmax or logits outputs~--~or just the predicted label) as well as a full or partial access to the training data. Note that in order to thwart a possible restriction concerning the access to the training set, \citep{papernot2017practical} proposes a way to train a substitute model by synthetically generating data labeled by the neural network under attack. 

\subsubsection{Specificity induced by embedded models}
\label{note_threat_model}
For our experiments, we do not consider a strict black-box setting since we assume an attacker will try to transfer the adversarial examples from one full-precision model to a quantified one or one quantified model to another one with a different level of quantization. This means that we assume a worst-case scenario where an adversary knows the model architecture and can query it without limitation with a full access to the softmax output. Moreover, since we use classical image collections, we assume that the attacker has access to the same dataset. Considering the global context of embedded neural networks for inference, numerous \textit{popular} and proven architectures (such as the ResNet networks) are directly applied for a large scope of applications. Then, a scenario where an adversary craft malicious inputs from a known full precision models to attack an optimized (i.e. quantized) model in, for example, a mobile device is a realistic scenario.

However, we must highlight an important characteristic of the threat models when dealing with an attacker who aims to target an embedded machine learning model. In that case, both the \textit{architecture} of the model itself \textit{and} its \textit{implementation} are important. That means we need to consider a twofold white/black box paradigm: on one hand, the adversary can have~--~classically~--~ full or no knowledge of the model architecture (abstraction level), and, on the other hand, he may also have full or no knowledge of the model implementation within the \textit{target device} (physical level).

As previously said, in this work and more particularly in the section dealing with the use of quantized networks as a defense mechanism, we mainly focus on a threat model where an attacker has a white-box access to the model architecture but not for its implementation in the target device. Then, the most natural scenario corresponds to an attacker that tries to directly transfer the adversarial examples crafted from a full precision model to the embedded system.

We are conscious of the limitation of such a scenario since, obviously, the attacker may guess~--~thanks to information about the hardware platform (i.e. memory, precision constraints, etc.)~--~relevant optimization methods applied to the model (weights and activations quantization, pruning...). That means an advanced adversary could try to craft adversarial examples from a quantized model of its own (without knowing the quantization method used for the target device neither if additional optimizations have been performed).


\section{Experiments}
\label{Experiments}
We start by performing adversarial robustness experiments for full-precision and quantized models with gradient-based and gradient-free attacks that may be used in black-box settings (unfeasible gradient computation).
In both case, we find that quantization does not provide reliable protection. We notice that quantization causes some gradient masking, which tampers some gradient-based attacks (FGSM and BIM) and may prevent gradient-free attacks relying on the approximation of the gradient of the output function (ZOO) to perform well. However, some gradient-based attacks using the STE to mount gradient-based attacks (CWl2) seem to avoid the gradient masking effect.
Secondly, we perform transferability experiences between full-precision and quantized models and show poor transferability capacities, which we explain with the \textit{quantization value shift} phenomenon and gradient misalignment.\\

\subsection{Data}
\label{Data}
        
We conduct our experiments on the CIFAR10\footnote{https://www.cs.toronto.edu/~kriz/cifar.html} and SVHN\footnote{http://ufldl.stanford.edu/housenumbers/} (Street View House Numbers)  two classical natural scene image datasets. CIFAR10 is composed of
60,000 images, with 10 classes. We use a training set of size 50,000 and a testing set of 10,000. The SVHN dataset is composed of 99,289 images, with 10 classes. We use a training set of size 73,257 and a testing test of 26,032.

    \subsection{Experience details}
    \label{Experience details}

For each dataset, we trained a full-precision (32-bit floating point) neural network (hereafter called "float model" in tables), and various quantized neural networks. More precisely, for each data set, the neural network architecture is based on the one presented in \citep{courbariaux2016binarized}. It consists of convolutional blocks, each of them being the stack of a convolution layer, a batch-normalization layer and the relu activation function, followed by dense blocks being a stack of a dense layer, a batch-normalization layer and the ReLu activation function. At the top of the network, we chose a dense layer with the softmax activation function, contrary to \citep{courbariaux2016binarized}, where there is no activation function but a final batch normalization layer. Models architecture are detailed in Appendix \ref{Network architecture}.
Full-precision and quantized networks were trained with the cross-entropy loss, contrary to \citep{courbariaux2016binarized} where the hinge-loss is used, as we found it to converge faster. The optimization is done with Adam \citep{kingma2014adam}, using a staircase decay for the learning rate.\\
Four different quantization bitwidth are considered: 1,2,3 and 4 bits. For each bitwidth, we consider quantization on the weights only (weight quantization) or the weights and the output of each convolutional or dense block (full quantization). For the weight binarization, the full binarization and the 2,3,4-bit quantization, we use respectively the Binary Connect method \citep{courbariaux2015binaryconnect}, the Binary Net method \citep{courbariaux2016binarized} and the Dorefa Net method \citep{zhou2016dorefa} described in \ref{Low bitwidth quantization}. The input layer and the last dense layer are never quantized, to allow an efficient training \citep{courbariaux2016binarized}. 

The performance (accuracy) of each model on the test sets is presented in Table \ref{model_accuracy_test}. Quantization does not affect significantly the accuracy, except for fully binarized models which achieve only 0.79 and 0.89 accuracy on CIFAR10 and SVHN respectively, which represents a non negligible drop of performance. For quantized models with more than 1 bit, the test set accuracy is comparable to the one obtained for full-precision models. These results are consistent with \citep{courbariaux2016binarized} and \citep{zhou2016dorefa}. Note that in \citep{courbariaux2016binarized} the authors explain the performance of binarized networks with a regularization effect brought by quantization and \citep{zhou2016dorefa} show that the architecture as well as the size of the data set can have an impact on the performance of quantized networks.\\

\begin{table}[h!]
\begin{center}
\begin{tabularx}{\linewidth}{|c|*{8}{>{\centering \arraybackslash}X|}}
\hline
& \multicolumn{4}{c|}{CIFAR10} & \multicolumn{4}{c|}{SVHN} \\ 
\hline
\textbf{Full-precision} & \multicolumn{4}{c|}{0.89} & \multicolumn{4}{c|}{0.96} \\
\hline
\textbf{Bitwidth} & \textbf{1} & \textbf{2} & \textbf{3} & \textbf{4} & \textbf{1} & \textbf{2} & \textbf{3} & \textbf{4} \\
\hline
\textit{Full quantization} & 0.79 & 0.87 & 0.88 & 0.88 & 0.89 & 0.95 & 0.95 & 0.95 \\
\hline
\textit{Weight quantization} & 0.88 & 0.88 & 0.88 & 0.88 & 0.96 & 0.95 & 0.96 & 0.95 \\
\hline
\end{tabularx}
\end{center}
\caption{Models accuracy on test sets. Full quantization means that both weights and activation values are quantized.}
\label{model_accuracy_test}
\end{table}

For each data set, we begin by evaluating the robustness of the full-precision and quantized models when the adversary uses three classical white-box gradient-based attacks: FGSM, BIM and CWl2. Then, we evaluate two gradient-free attacks, suitable for black-box settings: ZOO and SPSA. For FGSM, BIM, CWl2 and SPSA we use the Cleverhans library \citep{papernot2018cleverhans}, and for ZOO we use the original code provided by the authors\footnote{https://github.com/huanzhang12/ZOO-Attack}. The attacks parameters are detailed in Appendix \ref{Attack parameters}. BIM, CWl2, SPSA and ZOO are performed on 1000 randomly samples from the test set. 
\\
Over a second phase, we evaluate the transferability of attacks between full-precision and quantized models.\\

    \subsection{Evaluation metrics}
    \label{Evaluation metrics}

For $x \in \mathbb{R}^m, p \in \mathbb{N}, p \geq 1$, we note $l_p$ the p-norm of $x$:
\begin{equation*}
    \left\| x \right\|_p = \left( \sum_{i=1}^m | x_i |^p \right)^{\frac{1}{p}}
\end{equation*}

For each attack, we report two evaluation metrics (see \citep{carlini2019evaluating} for an extended review of the adversarial robustness evaluation):
\begin{itemize}[label=--]
    \item The adversarial accuracy, which is the accuracy of the model on adversarial examples (noted \textit{acc} in the result tables). The crafting method generates an adversarial example $x'$ from each input $x$ of the test set $X$. Hereafter we note $X'$ the adversarial test set on which is computed the adversarial accuracy. The higher the adversarial accuracy, the less the model is fooled by adversarial examples, i.e. the more the model is robust against the attack.
    \item  The average minimum-distance of the adversarial perturbation, i.e. in our case, the average $l_2$ norm and $l_{\infty}$ norm of the difference between clean and adversarial examples which succeed to fool the target model (simply noted $l_2$ and $l_{\infty}$ in the result tables). This quantifies the average distortion needed by the attacker to fool the model. 
\end{itemize}

\section{Results}
\label{Results}
\subsection{Robustness against gradient-based and gradient-free attacks}
\label{Robustness against gradient-based and gradient-free attacks}    
    
Results of direct attacks against fully quantized and weight-only quantized models are presented respectively in tables \ref{White-box weights and activations} and \ref{White-box weights only}. For these tables and the following ones dealing with quantized models, first row of the results is for 1-bit model (binarized model), second row for the 2-bit model, third row for the 3-bit model and fourth row for the 4-bit model. 

\begin{table}[h!]
\begin{center}
\scalebox{0.9}{
\begin{tabularx}{\linewidth}{ccccccccccccc}
\hline
\\~\vspace{-20pt}\\
\multirow{4}{*}{~} & \multicolumn{6}{c}{CIFAR10} & \multicolumn{6}{c}{SVHN}

\\~\vspace{-20pt}\\
\\&\multicolumn{3}{c}{Float model} & \multicolumn{3}{c}{Quantized models} & \multicolumn{3}{c}{Float model} & \multicolumn{3}{c}{Quantized models}

\\&\multicolumn{3}{c}{\textit{(32-bit)}} & \multicolumn{3}{c}{\textit{(1,2,3,4-bit)}} & \multicolumn{3}{c}{\textit{(32-bit)}} & \multicolumn{3}{c}{\textit{(1,2,3,4-bit)}}

\\~\vspace{-20pt}\\
\\& acc &  $l_2$ &  $l_\infty$ & acc & $l_2$ &  $l_\infty$ & acc & $l_2$ &  $l_\infty$ & acc & $l_2$ &  $l_\infty$ \\
\hline

\multirow{4}{*}{FGSM} & \multirow{4}{*}{0.12} & \multirow{4}{*}{1.65} & \multirow{4}{*}{ 0.03 } & \textbf{0.66} & 1.65& 0.03 & \multirow{4}{*}{0.29} & \multirow{4}{*}{ 1.66} & \multirow{4}{*}{ 0.03} & \textbf{0.78} & 1.64 & 0.03  \\
&&&& 0.19 & 1.65 & 0.03 &&&& 0.39 & 1.66 & 0.03 \\
&&&& 0.17 & 1.65 & 0.03  &&&& 0.37 & 1.66  & 0.03\\
&&&& 0.18 & 1.65 & 0.03 &&&& 0.4 & 1.66 & 0.03 \\
\hline 
\multirow{4}{*}{BIM} & \multirow{4}{*}{0.07} & \multirow{4}{*}{ 1.17 } & \multirow{4}{*}{0.03} & \textbf{0.66}&  1.01& 0.03  & \multirow{4}{*}{0.05 } & \multirow{4}{*}{1.16 } & \multirow{4}{*}{0.03} & \textbf{0.79}& 1.0 &0.03 \\  
&&&& 0.06 &1.14 &0.03   &&&& 0.11& 1.13 &0.03 \\
&&&& 0.11 & 1.17 &0.03  &&&& 0.11 &1.13 &0.03\\
&&&& 0.06 &1.14 &0.03   &&&& 0.1 &1.13& 0.03 \\
\hline 
\multirow{4}{*}{CWl2} & \multirow{4}{*}{0.03} & \multirow{4}{*}{ 0.58 } & \multirow{4}{*}{0.04} & \textbf{0.11} & 0.78 & 0.08 & \multirow{4}{*}{0.02 } & \multirow{4}{*}{0.64 } & \multirow{4}{*}{0.06} & \textbf{0.06} &  1.02 & 0.1  \\ 
&&&& 0.06 & 0.6 & 0.04 &&&& 0.03 & 0.67 &0.07\\
&&&& 0.09 & 0.55 & 0.04  &&&& 0.02 & 0.66 & 0.07\\
&&&& 0.05 & 0.6 & 0.04 &&&& 0.02 & 0.68 & 0.07\\
\hline 
\hline
\multirow{4}{*}{SPSA} & \multirow{4}{*}{0.0 } & \multirow{4}{*}{1.37 } & \multirow{4}{*}{0.03} & \textbf{0.16} & 1.31 & 0.03 & \multirow{4}{*}{0.01 } & \multirow{4}{*}{1.38 } & \multirow{4}{*}{ 0.03} & \textbf{0.4} & 1.32&  0.03  \\
&&&& 0.0 &1.34 &0.03 &&&& 0.14 &1.34 &0.03\\
&&&& 0.0 &1.36 &0.03 &&&& 0.07 &1.35 &0.03\\
&&&& 0.0 &1.36 &0.03 &&&& 0.04& 1.37 &0.03\\
\hline
\multirow{4}{*}{ZOO} & \multirow{4}{*}{0.0 } & \multirow{4}{*}{0.72 } & \multirow{4}{*}{0.09} & 0.56 &0.1 &0.05 & \multirow{4}{*}{0.0 } & \multirow{4}{*}{0.91 } & \multirow{4}{*}{0.11} & 0.82& 0.07 &0.05 \\  
&&&& \textbf{0.83}& 0.13 &0.06 &&&& 0.93 &0.1 &0.06\\
&&&& 0.76 &0.24 &0.07 &&&& \textbf{0.94} &0.11 &0.05\\
&&&& 0.73 &1.09& 0.14 &&&& 0.93 &0.38 &0.1\\
\hline
\end{tabularx}
}  
\end{center}
\caption{Adversarial accuracy and distortions for gradient-based and gradient-free attacks against full-precision (32-bit) and fully quantized models.}
\label{White-box weights and activations}
\end{table}

 \begin{table}[h!]
\begin{center}
\scalebox{0.9}{
\begin{tabularx}{\linewidth}{ccccccccccccc}
 \hline
 \\~\vspace{-20pt}\\
 \multirow{4}{*}{~} & \multicolumn{6}{c}{CIFAR10} & \multicolumn{6}{c}{SVHN}

 \\~\vspace{-20pt}\\
 \\&\multicolumn{3}{c}{Float model} & \multicolumn{3}{c}{Quantized models} & \multicolumn{3}{c}{Float model} & \multicolumn{3}{c}{Quantized models}

 \\&\multicolumn{3}{c}{\textit{(32-bit)}} & \multicolumn{3}{c}{\textit{(1,2,3,4-bit)}} & \multicolumn{3}{c}{\textit{(32-bit)}} & \multicolumn{3}{c}{\textit{(1,2,3,4-bit)}}

 \\~\vspace{-20pt}\\
 \\& acc &  $l_2$ &  $l_\infty$ & acc & $l_2$ &  $l_\infty$ & acc & $l_2$ &  $l_\infty$ & acc & $l_2$ &  $l_\infty$ \\
 \hline

  \multirow{4}{*}{FGSM} & \multirow{4}{*}{0.12} & \multirow{4}{*}{ 1.65 } & \multirow{4}{*}{0.03} & 0.11 & 1.65& 0.03 & \multirow{4}{*}{0.29 } & \multirow{4}{*}{1.66 } & \multirow{4}{*}{0.03} & 0.28& 1.66 &0.03 \\  
  &&&& 0.18& 1.65 &0.03 &&&& 0.38 &1.66 &0.03\\
  &&&& 0.18& 1.65 &0.03  &&&& \textbf{0.4} &1.66 &0.03\\
  &&&& \textbf{0.19} &1.65 &0.03 &&&& 0.39 &1.66 &0.03\\ 
  \hline 
  
  \multirow{4}{*}{BIM} & \multirow{4}{*}{0.07 } & \multirow{4}{*}{1.17 } & \multirow{4}{*}{0.03} & 0.07 &1.19& 0.03 & \multirow{4}{*}{0.05 } & \multirow{4}{*}{1.16 } & \multirow{4}{*}{0.03} & 0.07 &1.16 &0.03 \\  
  &&&&0.08 &1.15 &0.03  &&&&0.1 &1.14 &0.03\\
  &&&&0.06 &1.15 &0.03 &&&&\textbf{0.11} &1.14 &0.03\\
  &&&& 0.08 &1.15 &0.03  &&&& 0.09 &1.13 &0.03\\
  \hline 
  \multirow{4}{*}{CWl2} & \multirow{4}{*}{0.03 } & \multirow{4}{*}{0.58 } & \multirow{4}{*}{0.04} & 0.05 &0.57& 0.04 & \multirow{4}{*}{0.02 } & \multirow{4}{*}{0.64 } & \multirow{4}{*}{0.06} & 0.02& 0.64 &0.05  \\  
  &&&& \textbf{0.06} &0.6 &0.04  &&&& 0.02 &0.66 &0.06\\
  &&&& 0.05 &0.61 &0.04 &&&& 0.02 &0.67 &0.07\\
  &&&& \textbf{0.06} &0.62 &0.04 &&&& 0.02 &0.68 &0.07\\
  \hline 
  \hline 
  \multirow{4}{*}{SPSA} & \multirow{4}{*}{0.0 } & \multirow{4}{*}{1.37 } & \multirow{4}{*}{0.03} & 0.0 &1.38& 0.03 & \multirow{4}{*}{0.01 } & \multirow{4}{*}{1.38} & \multirow{4}{*}{ 0.03} & 0.01& 1.38 &0.03  \\  
  &&&&0.0 &1.37 &0.03 &&&& \textbf{0.04} &1.37 &0.03\\
  &&&& 0.0 &1.36 &0.03 &&&& \textbf{0.04} &1.37 &0.03\\
  &&&& 0.0 &1.36 &0.03 &&&& 0.03 &1.37 &0.03\\
  \hline
  \multirow{4}{*}{ZOO} & \multirow{4}{*}{0.0 } & \multirow{4}{*}{0.72} & \multirow{4}{*}{ 0.09} & 0.0 & 0.75 & 0.1 & \multirow{4}{*}{0.0 } & \multirow{4}{*}{0.91} & \multirow{4}{*}{ 0.11} & 0.0& 0.92 &0.1 \\  
  &&&& 0.0 &0.74 &0.1 &&&& 0.0 &0.92 &0.11\\
  &&&& 0.0 &0.72 &0.09 &&&& 0.0& 0.95 &0.11\\
  &&&& 0.0 &0.73 &0.09 &&&& 0.0 &0.93 &0.11\\
  \hline 
 \end{tabularx}
 }
\end{center}
\caption{Adversarial accuracy and distortions for gradient-based and gradient-free attacks against full-precision (32-bit) and weight-only quantized models.}
\label{White-box weights only}
\end{table}

    \subsubsection{Robustness of binarized neural networks}
    \label{Robustness of binarized neural networks}

A first observation from the comparison of table \ref{White-box weights and activations} and \ref{White-box weights only} is that the weight-only quantization has no impact on the robustness. Then, with
table \ref{White-box weights and activations} we see that fully binarized models are far more robust to FGSM and BIM than their full-precision counterparts, as noted by \citep{galloway2017attacking}, but achieve only 0.79 and 0.89 accuracy on (respectively) the CIFAR10 and SVHN test datasets (see table \ref{model_accuracy_test}) which represents a non negligible drop of performance. 

However, CWl2~--~one of the most powerful crafting method~--~is almost as efficient against fully binarized neural networks as against full-precision models. Therefore, fully-binarized neural networks do not bring much robustness improvement compared to full-precison models against gradient-based attacks, as claimed in \citep{galloway2017attacking}.
We also note that fully binarized models are relatively more robust to SPSA as well compared to full-precision models. This combined with the slightly poorer performance of CWl2 on binarized models indicates that the loss surface for binarized models is difficult to optimize over.

For full quantization with more than 1 bit, the gradient-based attacks are almost as efficient as against a full-precision model, except for FGSM on SVHN only with a 10\% gain of accuracy.\\

    \subsubsection{Activation quantization causes gradient masking}

Interestingly, we see that ZOO fails very often to produce adversarial examples when attacking fully quantized neural networks. More precisely, we note that when adversarial examples are crafted on a full-precision model, ZOO and CWl2 reach almost the same adversarial accuracy, with slightly higher $l_2$ distortion for ZOO. However, when adversarial examples are crafted on a model with quantized weights and activations, we note that the adversarial accuracy is higher with ZOO than with CWl2, but that successful adversarial examples crafted with ZOO have much lower $l_2$ distortion than the ones crafted with CWl2 as well as the one observed for the full-precision model ($l_2=0.72$).
We claim that these observations reveal some form of gradient masking caused by the quantization of activation values. Firstly, the almost equal performance of ZOO compared to CWl2 for a full-precision model is expected as gradient-free attacks are supposed to perform worse than their gradient-based counterparts when no gradient masking occurs. Secondly, we argue that the phenomenon observed on full quantization models is due to gradient masking and the STE technique involved in these models training. 

We explain this by distinguishing two cases:
\begin{itemize}
    \item[--] ZOO fails to produce successful adversarial examples and CWl2 succeeds: (1) because of the activation quantization, a little change ($h e_i$) may switch the activation value from one quantization bucket to another, inducing a big change in the predicted softmax values, causing the discrete derivative $F^d_{ZOO}(x)$ to explode; (2) on the contrary, this change can also have no impact (keep values in the same bucket), which results in $F_{ZOO}(x - he_i, y) = F_{ZOO}(x + he_i, y)$, causing the discrete derivative $F^d_{ZOO}(x)$ to be null. To sum up, $F_{ZOO}$ presents some sharp curvatures or flatness around some points, caused by activation quantization, which prevents ZOO to build successful adversarial examples. The CWl2 attack avoids this problem as it computes gradients thanks to a STE, even if the gradient computed may not be exactly the same as the true gradient \citep{khalil2018combinatorial}.
    \item[--] Both ZOO and CWl2 succeed to produce successful adversarial examples: the $l_2$ distortion for the successful adversarial examples produced by ZOO is smaller than the one produced by CWl2. Around these points the surface of the objective function to optimize does not present any sharp curvature or flatness. Both ZOO and CWl2 do not suffer from the local minima problem, but as noted by \citep{khalil2018combinatorial}, the gradients computed by the CWl2 attack is not representative of the true gradient. The gradient being better estimated by the ZOO attack, it explains its success (lower $l_2$ distortion).
\end{itemize}

The gradient masking phenomenon hypothesis concerning the quantization of activation values is also verified with the fact that the SPSA attack (a gradient-free attack) performs quite better in terms of adversarial accuracy than the BIM attack, against fully binarized models. We also make the hypothesis that SPSA avoids the sharp curvatures or flatness observed around some points, where ZOO fails to produce adversarial examples, because of the more efficient gradient estimation method suited to noisy objective functions \citep{uesato2018adversarial}. 

For the weight quantized models results presented in Table \ref{White-box weights only}, we do not observe the same phenomena as for the full quantization models. No apparent robustness is noticeable for the weight quantized models. No sharp variation or flatness is induced for the objective function of the ZOO attack of the weight quantized models, as it originated from the quantization of activation values. It has to be noted that we also measured that the variance of the logits values between full-precision and weight-only quantized models is almost the same.

\subsection{Transferability}
\label{Transferability}
\begin{figure}[h]
	\centering
	\includegraphics[width=1\textwidth]{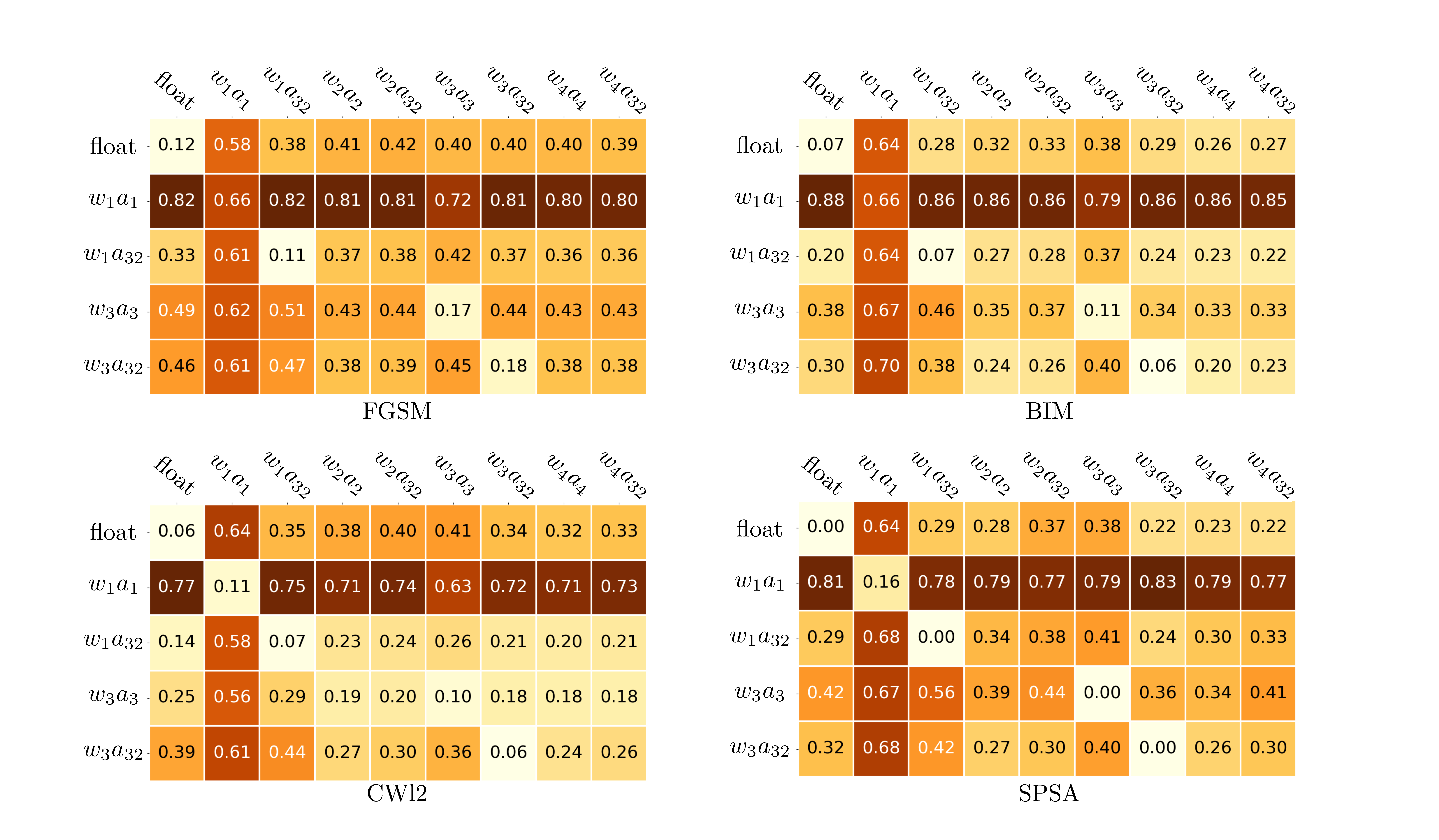} 
	\caption{Adversarial transferability results for CIFAR10. Rows are relative to source networks and columns to target networks. Values correspond to adversarial accuracy. The lower the value, the more transferability occurs.}
    \label{heatmap_cifar10}
\end{figure}

\begin{figure}[h]
	\centering
	\includegraphics[width=1\textwidth]{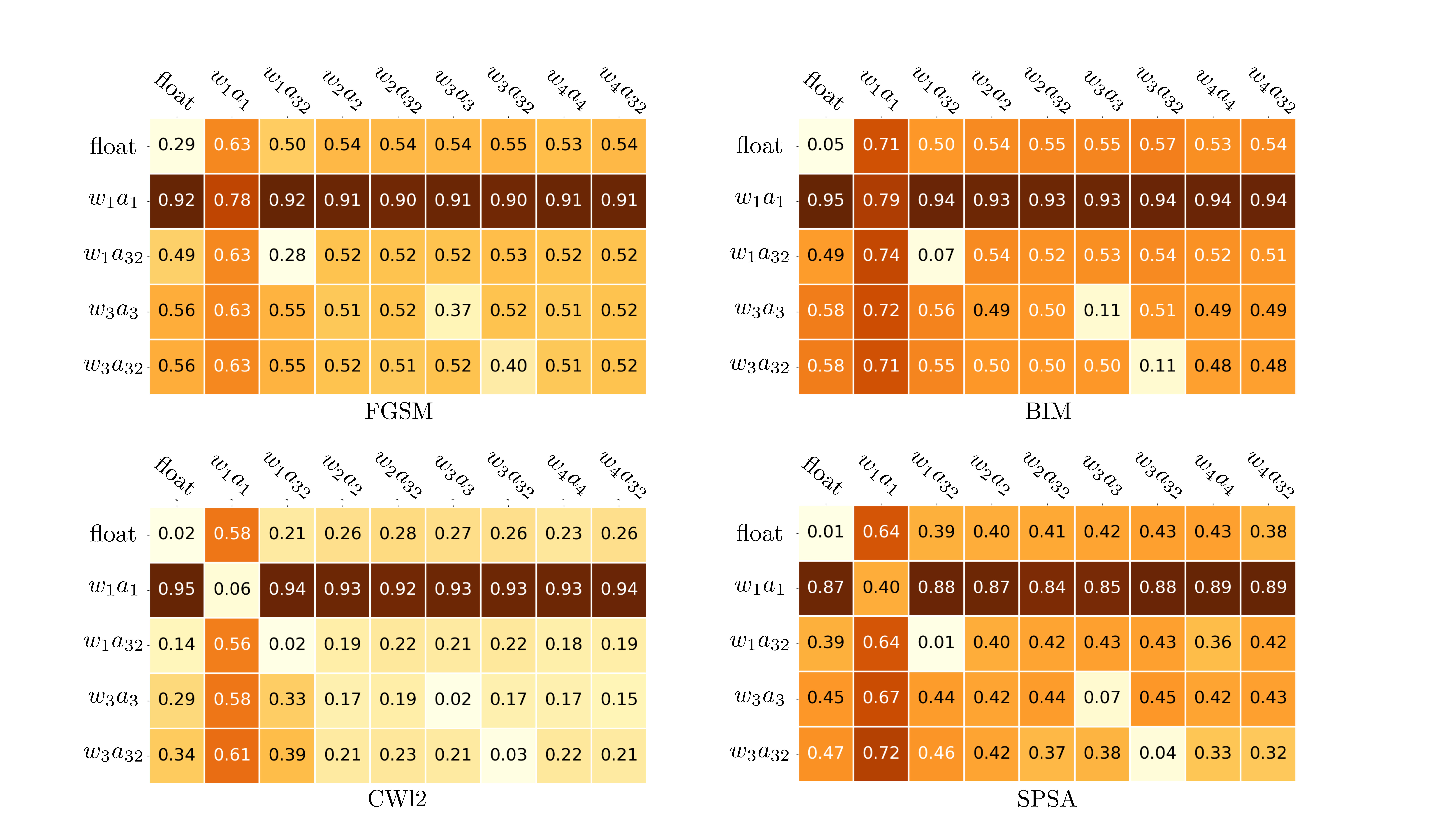}  
	\caption{Adversarial transferability results for SVHN. 
	Rows are relative to source network and columns to target networks. Values correspond to adversarial accuracy. The lower the value, the more transferability occurs.}
    \label{heatmap_svhn}
\end{figure}

We present results of transferability when the source network (i.e. the models the adversarial examples are crafted from) is full-precision, fully or weight-only quantized models, and the adversarial examples are transferred to (target models) full-precision, fully or weight-only quantized networks, in figures \ref{heatmap_cifar10} and \ref{heatmap_svhn}. For CWl2, as advised in \citep{carlini2017towards}, we consider $\kappa > 0$ to build strong adversarial examples on the source model, more likely to transfer. We tested $\kappa = 5, 10, 15, 20, 25$. For each source model, we report the best transferability results. The results for the cases where the source networks are 2-bit or 4-bit quantized models are not presented here for paper length purpose and as these results can be interpolated from the one we present. More complete tables can be found in Appendix \ref{Black-box other results}.\\

    \subsubsection{Weak transferability}
 
A first observation is that transferability results are quite poor for FGSM, BIM and SPSA. CWl2, given tuning the parameter $\kappa$, suffers less from transferability issues, at the cost of increased $l_2$ and $l_{\infty}$ distortion, except when the source or target network is a fully binarized network. Indeed, for the $\kappa$ values tested (for fully binarized models, the results reported are for $\kappa = 5$), when the source network is a fully binarized model, 
CWl2 struggles to find adversarial examples having both $F(x+\epsilon,y) < 0$ (see Equation \ref{eq_cwl2}) and a little $l_2$ distortion. This results in adversarial examples being missclassified but not imperceptible by a human. We hypothesize this comes from the hard to optimize loss function as noted in \ref{Robustness of binarized neural networks}. 
We also note that, as already noticed by \citep{wu2018understanding}, and contrary to what was initially found by \citep{kurakin2016adversarial2}, that BIM~--~as it is the case here~--~may produce more transferable adversarial examples than FGSM . 

    \subsubsection{Quantization shift phenomenon}

\begin{figure}[h]
	\centering
	\includegraphics[width=0.75\textwidth]{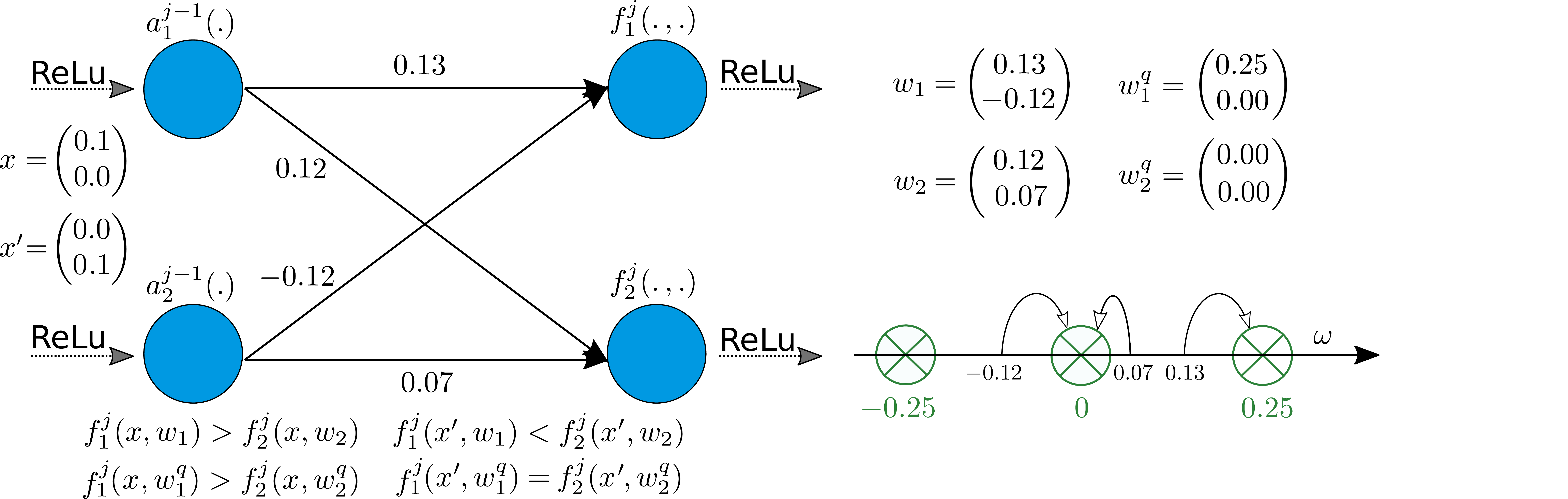}
	\caption{A toy example to illustrate the \textit{quantization value shift} phenomenon. Quantization of the weights cancels the adversarial effect.}
    \label{fig_quant_effect}
\end{figure}

These poor transferability results (mainly for FGSM and BIM) can be explained by the \textit{quantization value shift} phenomenon which takes places when quantization ruins the adversarial effect by mapping two different values to the same quantization bucket. In case of activation quantization, two activation values can be mapped to the same value. In case of weight quantization, this levelling effect may also be observed and ruins the adversarial effect. The Figure \ref{fig_quant_effect} shows a toy example of the impact of weight quantization on adversarial effect: in this example, the adversarial effect is canceled.\\

Consequently, whatever the quantization level of the source model adversarial examples are crafted on, evaluating them on a target model with a different quantization level may hinder their efficiency because of this phenomenon. 




    \subsubsection{Gradient misalignment}

\begin{figure}[h]
	\centering
	\includegraphics[width=0.75\textwidth]{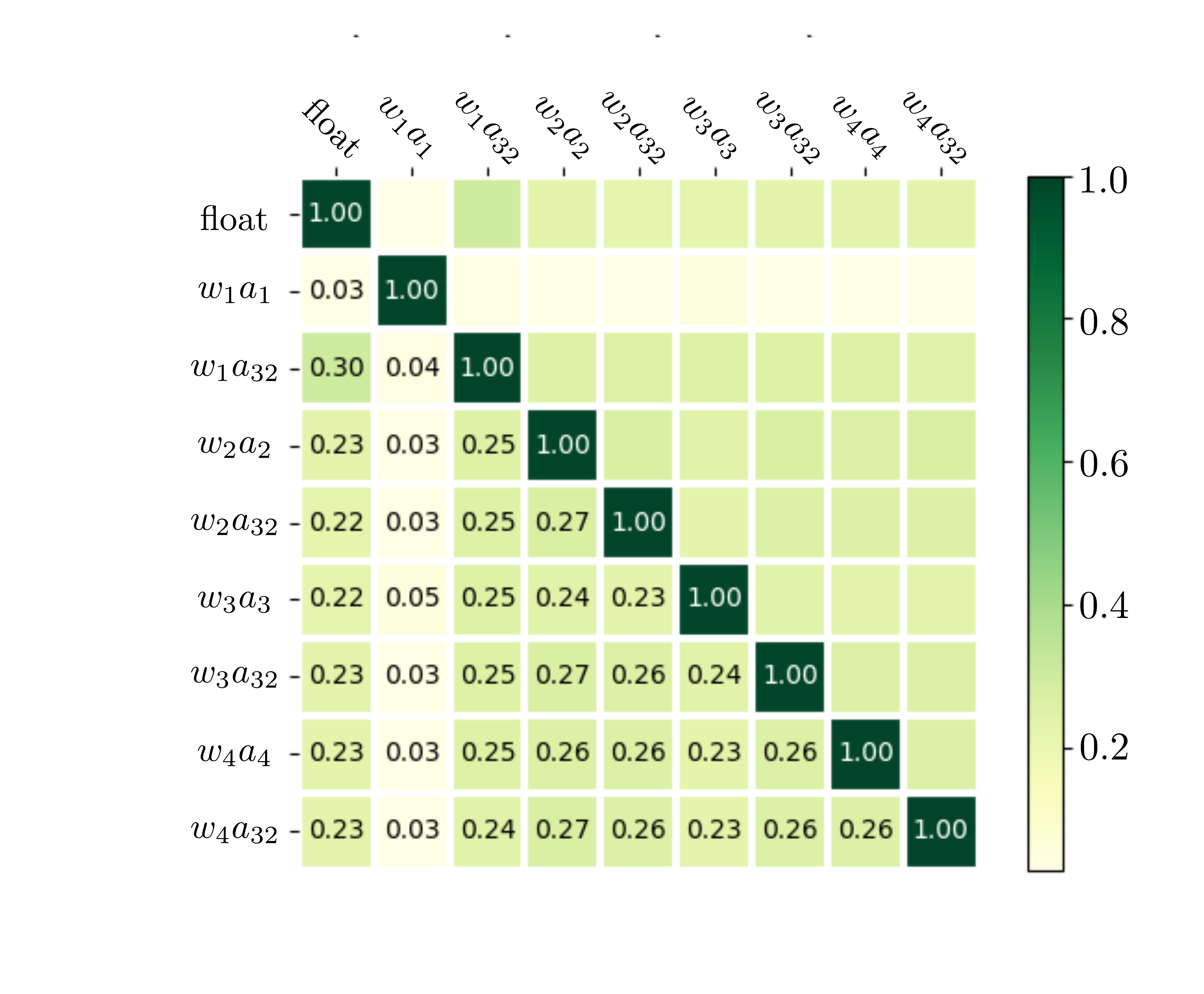}
	\caption{Cosine similarity values between the gradient of the loss function with respect to the input, for full-precision and quantized models. $w_ia_j$ designates a model with a \textit{i}-bit weight quantization and a \textit{j}-bit activation quantization.}
    \label{cos_sim_results}
\end{figure}

Regarding the transferability results, we may also hypothesize that the gradient direction between float models and quantized models and between models with different bitwidths is quite different. This gradient misalignment may be noticeable for the gradient computed with the \textit{Straight Through Estimator}, as poor transferability is observed for the white-box attacks (FGSM, BIM, CWl2), and for the real gradient, as poor transferability is observed for SPSA. We measure the mean cosine similarity between the gradient of the loss function with respect to the input between models with different bitwidth and show the results in Figure \ref{cos_sim_results} for CIFAR10.
We remind that the cosine similarity $CS(a,b) \in \mathbb{R}$ between two vectors $a,b \in \mathbb{R}^m \times \mathbb{R}^m$ is defined as:\\
\begin{equation*}
    CS(a,b) = \frac{\langle {a,b} \rangle}{\left\| a \right\|_2 \cdot \left\| b \right\|_2}
\end{equation*}
where $\langle {\cdot,\cdot}\rangle$ is the usual scalar product. $CS(a,b)=0$ indicates orthogonal vectors for the usual scalar product, $CS(a,b)=1$ indicates aligned vectors in the same direction and $CS(a,b)=-1$ indicates aligned vectors in opposite directions.\\

In Figure \ref{cos_sim_results}, we first observe that the cosine similarity values for gradients of the loss function between full-precision and quantized models and between quantized models with different bitwidths, are relatively close to 0, indicating nearly orthogonal gradient directions. We observe that the cosine similarity for the gradient of the loss function with respect to the input between fully-binarized models and others models is the closest to 0.  This is in accordance for example with results presented in Fig. \ref{heatmap_cifar10} where transferability capacities for FGSM, BIM, CWl2 and SPSA are the poorest when fully-binarized models are involved. Moreover, this may explain the fact that adversarial accuracy is much higher in tables \ref{w1 a1 to fully} and \ref{w1 a1 to weight} (Appendix \ref{Black-box other results}), where adversarial examples are crafted on  fully-binarized models, than in the other tables
(see Appendix \ref{Black-box other results}).


To conclude, transferability results show that quantization strongly alters chances of success for an adversary who has only access to a full-precision (or quantized) version of a model and wants to attack a quantized (respectively, a full-precision) version of a model, assuming this adversary can not use a black-box attack such as the SPSA one.\\

\section{Ensemble of quantized models}
\label{Ensemble of quantized models}
\subsection{Motivation}
\label{Motivation_2}
Regarding of the transferability results, a logical consequence and a natural assumption is to consider an ensemble of quantized models to filter out adversarial examples. In this section, we analyze the relevance of this defense strategy. We remind the reader about the important point we highlight in section \ref{note_threat_model} about the threat models and their intrinsic limitations.

We consider an ensemble of quantized models, $\mathcal{M}=\{M_i\}$, in our case: a full-precision model and four fully-quantized models (1,2,3 and 4 bits). We first analyze statistically how the models agree on clean and adversarial test set using different crafting methods (FGSM, BIM, CWl2 and SPSA). More precisely, we consider an adversarial example crafted on a source model $M_{s}$ and we look at how the five models agree, given that this adversarial example is successful or not on $M_{s}$. Similarly, we look at how the five models agree on clean test set examples, given that this test set example is correctly classified in the true class or, on the contrary, misclassified  by $M_s$.
Our main observations are the following:
\begin{enumerate}
    \item the models are more likely to agree on clean samples than on adversarial examples (successful or not);
    \item the models are much more likely to agree on unsuccessful (on $M_{s}$) adversarial examples. 
\end{enumerate}

In table \ref{transfer_result_example}, we show for CIFAR10 and SVHN, given the source model $M_s$, how the four other models agree on test set and adversarial examples crafted with FGSM. For example, for CIFAR10, with the 2-bitwidth model, 77\% of test set examples well-recognized by $M_s$ are also correctly classified by the other four models, compared to 33\% for misclassified test set examples. Moreover, 37\% of successful adversarial examples (i.e. examples that effectively fool $M_s$) also fool the other four models, and 76\% of unsuccessful adversarial examples (on $M_s$) are also unsuccessful on the other four models.\\
\begin{table}[h!]
\begin{center}
\begin{tabularx}{\linewidth}{XXXXXXX||XXXXX|}
\hline
\multicolumn{2}{|c|}{} & \multicolumn{5}{c}{CIFAR10} & \multicolumn{5}{c|}{SVHN}\\
\multicolumn{2}{|c|}{} & \multicolumn{10}{c|}{\textbf{Source model}}\\
\cline{3-12}
\multicolumn{2}{|c|}{} & $float$ & $w_1 a_1$ & $w_2 a_2$ & $w_3 a_3$ & $w_4 a_4$
& $float$ & $w_1 a_1$ & $w_2 a_2$ & $w_3 a_3$ & $w_4 a_4$ \\ 
\cline{3-12}
\multicolumn{2}{|c|}{} & \multicolumn{10}{c|}{Test set examples}\\
\hline
\multicolumn{2}{|c|}{\textit{Correctly classified}} & 0.75 & 0.85 & 0.77 & 0.82 & 0.76 
& 0.9 & 0.97 & 0.91 & 0.91 & 0.9\\
\multicolumn{2}{|c|}{\textit{Misclassified}} & 0.39 & 0.19 & 0.33 & 0.33 & 0.34
& 0.46 & 0.16 & 0.38 & 0.37 & 0.4\\
\hline
\multicolumn{2}{|c|}{} & \multicolumn{10}{c|}{Adversarial examples (FGSM)}\\
\hline
\multicolumn{2}{|c|}{\textit{Successful}} & 0.31 & 0.09 & 0.37 & 0.31 & 0.33 
& 0.33 & 0.09 & 0.41 & 0.43 & 0.39\\
\multicolumn{2}{|c|}{\textit{Unsuccessful}} & 0.88 & 0.80 & 0.76 & 0.78 & 0.79 
& 0.89 & 0.94 & 0.81 & 0.82 & 0.84\\
\hline
\end{tabularx}
\end{center}
\caption{Rate of examples for which the other four models agree, depending on the source model $M_s$ and its prediction results (\textit{correctly classified}, \textit{misclassified}, \textit{successful}, \textit{unsucessful}).}
\label{transfer_result_example}
\end{table}

We design the following ensemble-based defense method for the distant system: the prediction for an upcoming example is done only if $m$ or more models agree, and the final label is the one predicted by these $m$ models. 
We note $valid_{m, \mathcal{M}}(X)$ the set of samples from the input data set $X$ which respect this criterion:
\begin{equation}
    valid_{m, \mathcal{M}}(X) = \left\{ x \in X \left | \
    \max_{j \in \left[ 1,\dots,C \right]} \ 
    \sum_{i} \textbf{1}_{M_i (x) = j} \ge m \right. \right\}   
\end{equation}
with $C$ the number of labels and $M_{i}\left( x \right)$ the output prediction label of $x$ by $M_{i}$. Then, the \textit{prediction rate} (hereafter, PR) quantifying the proportion of examples from $X$ for which prediction is performed is :
\begin{equation}
    PR_{m, \mathcal{M}}(X) = \frac{\left|valid_{m, \mathcal{M}}(X)\right|}{\left|X\right|}
\end{equation}
Where $\left|X\right|$ is the cardinality of $X$. Considering the remarks made above, this approach encourages prediction for clean examples rather than for adversarial examples, and when the prediction is performed on adversarial examples, it would be predominantly unsuccessful ones. 

When evaluating this defense on the adversarial test set ($X'$), an overall performance metric, hereafter called \textit{defense accuracy} (d\_\text{acc}) is simply defined as the proportion of adversarial examples which have been filtered out or which are unsuccessful. Practically, the defense accuracy could also be defined as:  

\begin{equation}
    d\_\text{acc}_{m, \mathcal{M}}(X') = 1- \frac{\left|valid_{m, \mathcal{M}}(X')^{s}\right|}{\left|X'\right|}
    \label{error_rate_ensemble}
\end{equation}

Where $valid_{m, \mathcal{M}}(X')^{s}$ denotes the set of successful adversarial examples which thwart the filtering process (i.e. the \textit{error rate} of the defense). 

Thus, the number $m$ has to be decided following a trade-off between the number of test set examples for which prediction is performed (which has to remain high) and the defense error rates (which has to be low) when facing adversarial examples. We experimentally set $m=4$ for CIFAR10 and $m=5$ for SVHN to reach a good trade-off. As presented in Table \ref{ensemble_cifar10_svhn_test}, the prediction is performed for more than 87\% of the clean test set examples for both CIFAR10 and SVHN, with an accuracy of 90\% and 98\% respectively for CIFAR10 and SVHN on clean test set examples for which prediction is performed. 

\begin{table}[h!]
\begin{center}
\begin{tabularx}{\linewidth}{c*{4}{>{\centering \arraybackslash}X}}
 \hline
 \\~\vspace{-20pt}\\
 \multirow{4}{*}{~} & \multicolumn{2}{c}{CIFAR10} & \multicolumn{2}{c}{SVHN}
\\~\vspace{-20pt}\\
 \\& PR & accuracy & PR & accuracy \\
 \hline
 \\~\vspace{-20pt}\\
 Test set & 0.87 & 0.90 & 0.87 & 0.98
 \end{tabularx}
\end{center}
\caption{Prediction rate and accuracy with the ensemble of models on CIFAR10 and SVHN.}
\label{ensemble_cifar10_svhn_test}
\end{table}

\subsection{Results}
\label{Results_2}

We present the results of this defense on CIFAR10 and SVHN in table \ref{ensemble_cifar10_svhn}. For each source model and each attack method, we report the defense accuracy $d\_\text{acc}$ against four attacks (FGSM, BIM, CWl2 and SPSA), along with the prediction rate PR.

\begin{table}[h!]
\begin{center}
\begin{tabularx}{\linewidth}{c*{8}{>{\centering \arraybackslash}X}}
 \hline
 \\~\vspace{-20pt}\\
 \multirow{4}{*}{~} & \multicolumn{4}{c}{CIFAR10} & \multicolumn{4}{c}{SVHN}

 \\~\vspace{-20pt}\\
 \\&\multicolumn{2}{c}{Float model} & \multicolumn{2}{c}{Quantized models} & \multicolumn{2}{c}{Float model} & \multicolumn{2}{c}{Quantized models}

 \\&\multicolumn{2}{c}{\textit{(32-bit)}} & \multicolumn{2}{c}{\textit{(1,2,3,4-bit)}} & \multicolumn{2}{c}{\textit{(32-bit)}} & \multicolumn{2}{c}{\textit{(1,2,3,4-bit)}}

 \\~\vspace{-20pt}\\
 \\& PR & d\_\text{acc} & PR & d\_\text{acc} & PR & d\_\text{acc} & PR & d\_\text{acc} \\
 \hline
 \multirow{4}{*}{FGSM} & \multirow{4}{*}{0.58} & \multirow{4}{*}{0.63} & 0.73 & \textbf{0.9} & \multirow{4}{*}{0.39} & \multirow{4}{*}{0.65} & 0.80 & \textbf{0.98}  \\  
 &&& 0.58 & 0.53  &&& 0.47 & 0.86   \\
 &&& 0.45 & 0.63   &&& 0.47 & 0.84  \\
 &&& 0.53 & 0.57  &&& 0.46 & 0.87 \\
 \hline
 \multirow{4}{*}{BIM} & \multirow{4}{*}{0.65} & \multirow{4}{*}{0.44} & 0.71 & \textbf{0.88} & \multirow{4}{*}{0.20} & \multirow{4}{*}{0.85} & 0.80 & \textbf{0.99} \\  
 &&& 0.62 & 0.38  &&& 0.28 & 0.81   \\
 &&& 0.57 & 0.44   &&& 0.26 & 0.81  \\
 &&& 0.52 & 0.48  &&& 0.25 & 0.82 \\
 \hline
\multirow{4}{*}{CWl2} & \multirow{4}{*}{0.54} & \multirow{4}{*}{0.60} & 0.17 & \textbf{0.84} & \multirow{4}{*}{0.15} & \multirow{4}{*}{0.84} & 0.18 & \textbf{0.84}  \\  
 &&& 0.47 & 0.53  &&& 0.23 & 0.77   \\
 &&& 0.58 & 0.42   &&& 0.2 & 0.8  \\
 &&& 0.36 & 0.64  &&& 0.17 & 0.83 \\
 \hline
 \multirow{4}{*}{SPSA} & \multirow{4}{*}{0.68} & \multirow{4}{*}{0.41} & 0.32 & \textbf{0.82} & \multirow{4}{*}{0.22} & \multirow{4}{*}{0.8} & 0.5 & \textbf{0.97}  \\  
 &&& 0.48 & 0.54  &&& 0.32 & 0.82   \\
 &&& 0.44 & 0.57   &&& 0.29 & 0.79  \\
 &&& 0.58 & 0.42  &&& 0.31 & 0.75 \\
 \hline
 
\end{tabularx}
\end{center}
\caption{Defense accuracy (d\_\text{acc}) and adversarial prediction rate (PR) for gradient-based and gradient-free attacks against an ensemble of quantized models, depending of the source model.}
\label{ensemble_cifar10_svhn}
\end{table}

Considering the natural threat model discussed in \ref{Threat model}, results are interesting for adversarial examples crafted from the full precision model especially for SVHN. If an attacker tries to directly transfer the adversarial examples crafted from the full precision model with, for example, the CWl2 attack, 60\% of the adversarial examples are harmless (i.e. filtered out or unsuccessful) for CIFAR10 and this robustness is even stronger for SVHN with a defense accuracy superior to 80\% for BIM, CWl2 or SPSA.

Moreover, coherently with the transferability results (see Figures \ref{heatmap_cifar10}, \ref{heatmap_svhn} and additional tables in Appendix \ref{Black-box other results}), the highest robustness is reached when adversarial examples are crafted from a fully binarized network with a defense accuracy superior to 0.8~--~whatever the crafting method~--~ particularly for SVHN. However, there is no significant gain compare to the transferability results obtained by taking each quantized model separately (see figures \ref{heatmap_cifar10} and \ref{heatmap_svhn}). But, except for this case of the fully binarized network, the ensemble of quantized models shows better robustness to transferred adversarial examples than all single models. The more relevant gain is reached with CWl2 attack with a mean (over the 2, 3 and 4-bitwidth networks) defense accuracy of 0.53 and 0.8 respectively for CIFAR10 and SVHN. 

Once again, if we do not claim to meet state-of-art detection based protection (such as \citep{meng2017magnet}, \citep{ilyas2017robust}, \citep{lu2017safetynet} and \citep{zheng2018robust}), we regard these results as significant ones, particularly since we are deeply convinced that an efficient defense strategy against adversarial examples will necessary be a composition of several protection schemes as it the case in other security domains such as efficient countermeasures against physical attacks for cryptographic systems which combine masking, hiding and redundancy principles.   

\section{Conclusion}
In this article, we show experimentally on CIFAR10 and SVHN and state-of-the-art gradient-based and gradient-free attacks that quantization in itself offers very poor protection against adversarial examples crafted by adversaries having access to the model or able to query it. 
We find that activation quantization can lead to gradient masking. We verify experimentally that the efficiency of some gradient-based and gradient-free attacks can thus be tampered but other gradient-based or gradient-free attacks do not suffer from gradient masking, because of the usage of a STE to approximate gradients, or the optimization procedure begin well-suited for noisy functions.
Eventually, we demonstrate poor transferability capacities between classical models and quantized models, and between quantized models with different bitwidths. We explain this by the \textit{quantization shift phenomenon} which ruins adversarial effects and gradients misalignment.

As an exploratory work and logical consequence of the transferability results, we analyze the impact of considering an ensemble of quantized models in order to filter out adversarial examples with a minimum impact on the natural accuracy. Such an ensemble method, like any other detection-based approach, suffers from a narrow threat model since the defense is useless with an attacker aware of the implementation details of the model in the target device \citep{carlini2017adversarial}. However, for black-box paradigms, the use of quantized ensemble may have an interesting impact on the transferability when associated to other and complementary defense mechanisms. 

As an important outcome of these experiments, we believe that the characteristics of embedded models particularly induced by quantization approaches (weights or activation outputs) have to be taken into consideration in order to design suitable and efficient protection schemes. These defense strategies for embedded models will be the purpose of future works,
since robustness requirements will obviously become more and more compulsory as critical tasks (as well as processed data) will be performed thanks to a growing variety of devices. 


\clearpage
\newpage


\bibliographystyle{ieeetr}
\bibliography{Biblio.bib} 

\begin{thebibliography}{10}

\bibitem{denil2013predicting}
M.~Denil, B.~Shakibi, L.~Dinh, N.~De~Freitas, {\em et~al.}, ``Predicting
  parameters in deep learning,'' in {\em Advances in neural information
  processing systems}, pp.~2148--2156, 2013.

\bibitem{hacene2018quantized}
G.~B. Hacene, V.~Gripon, M.~Arzel, N.~Farrugia, and Y.~Bengio, ``Quantized
  guided pruning for efficient hardware implementations of convolutional neural
  networks,'' {\em arXiv preprint arXiv:1812.11337}, 2018.

\bibitem{gong2014compressing}
Y.~Gong, L.~Liu, M.~Yang, and L.~Bourdev, ``Compressing deep convolutional
  networks using vector quantization,'' {\em arXiv preprint arXiv:1412.6115},
  2014.

\bibitem{han2015deep}
S.~Han, H.~Mao, and W.~J. Dally, ``Deep compression: Compressing deep neural
  networks with pruning, trained quantization and huffman coding,'' {\em arXiv
  preprint arXiv:1510.00149}, 2015.

\bibitem{choi2016towards}
Y.~Choi, M.~El-Khamy, and J.~Lee, ``Towards the limit of network
  quantization,'' {\em arXiv preprint arXiv:1612.01543}, 2016.

\bibitem{courbariaux2015binaryconnect}
M.~Courbariaux, Y.~Bengio, and J.-P. David, ``Binaryconnect: Training deep
  neural networks with binary weights during propagations,'' in {\em Advances
  in neural information processing systems}, pp.~3123--3131, 2015.

\bibitem{courbariaux2016binarized}
M.~Courbariaux, I.~Hubara, D.~Soudry, R.~El-Yaniv, and Y.~Bengio, ``Binarized
  neural networks: Training deep neural networks with weights and activations
  constrained to+ 1 or-1,'' {\em arXiv preprint arXiv:1602.02830}, 2016.

\bibitem{hubara2017quantized}
I.~Hubara, M.~Courbariaux, D.~Soudry, R.~El-Yaniv, and Y.~Bengio, ``Quantized
  neural networks: Training neural networks with low precision weights and
  activations.,'' {\em Journal of Machine Learning Research}, vol.~18, no.~187,
  pp.~1--30, 2017.

\bibitem{rastegari2016xnor}
M.~Rastegari, V.~Ordonez, J.~Redmon, and A.~Farhadi, ``Xnor-net: Imagenet
  classification using binary convolutional neural networks,'' in {\em European
  Conference on Computer Vision}, pp.~525--542, Springer, 2016.

\bibitem{li2016ternary}
F.~Li, B.~Zhang, and B.~Liu, ``Ternary weight networks,'' 2016.

\bibitem{zhu2016trained}
C.~Zhu, S.~Han, H.~Mao, and W.~J. Dally, ``Trained ternary quantization,'' {\em
  arXiv preprint arXiv:1612.01064}, 2016.

\bibitem{gupta2015deep}
S.~Gupta, A.~Agrawal, K.~Gopalakrishnan, and P.~Narayanan, ``Deep learning with
  limited numerical precision,'' in {\em International Conference on Machine
  Learning}, pp.~1737--1746, 2015.

\bibitem{zhou2016dorefa}
S.~Zhou, Y.~Wu, Z.~Ni, X.~Zhou, H.~Wen, and Y.~Zou, ``Dorefa-net: Training low
  bitwidth convolutional neural networks with low bitwidth gradients,'' {\em
  arXiv preprint arXiv:1606.06160}, 2016.

\bibitem{ding2017lightnn}
R.~Ding, Z.~Liu, R.~Shi, D.~Marculescu, and R.~Blanton, ``Lightnn: Filling the
  gap between conventional deep neural networks and binarized networks,'' in
  {\em Proceedings of the on Great Lakes Symposium on VLSI 2017}, pp.~35--40,
  ACM, 2017.

\bibitem{polino2018model}
A.~Polino, R.~Pascanu, and D.~Alistarh, ``Model compression via distillation
  and quantization,'' {\em arXiv preprint arXiv:1802.05668}, 2018.

\bibitem{szegedy2013intriguing}
C.~Szegedy, W.~Zaremba, I.~Sutskever, J.~Bruna, D.~Erhan, I.~Goodfellow, and
  R.~Fergus, ``Intriguing properties of neural networks,'' in {\em
  International Conference on Learning Representations}, 2013.

\bibitem{goodfellow2015laceyella}
I.~Goodfellow, J.~Shlens, and C.~Szegedy, ``Explaining and harnessing
  adversarial examples,'' in {\em International Conference on Learning
  Representations}, 2015.

\bibitem{carlini2017towards}
N.~Carlini and D.~Wagner, ``Towards evaluating the robustness of neural
  networks,'' in {\em 2017 IEEE Symposium on Security and Privacy (SP)},
  pp.~39--57, IEEE, 2017.

\bibitem{moosavi2016deepfool}
S.-M. Moosavi-Dezfooli, A.~Fawzi, and P.~Frossard, ``Deepfool: a simple and
  accurate method to fool deep neural networks,'' in {\em Proceedings of the
  IEEE Conference on Computer Vision and Pattern Recognition}, pp.~2574--2582,
  2016.

\bibitem{kurakin2016adversarial}
A.~Kurakin, I.~Goodfellow, and S.~Bengio, ``Adversarial machine learning at
  scale,'' {\em arXiv preprint arXiv:1611.01236}, 2016.

\bibitem{papernot2017practical}
N.~Papernot, P.~McDaniel, I.~Goodfellow, S.~Jha, Z.~B. Celik, and A.~Swami,
  ``Practical black-box attacks against machine learning,'' in {\em Proceedings
  of the 2017 ACM on Asia Conference on Computer and Communications Security},
  pp.~506--519, ACM, 2017.

\bibitem{chen2017zoo}
P.-Y. Chen, H.~Zhang, Y.~Sharma, J.~Yi, and C.-J. Hsieh, ``Zoo: Zeroth order
  optimization based black-box attacks to deep neural networks without training
  substitute models,'' in {\em Proceedings of the 10th ACM Workshop on
  Artificial Intelligence and Security}, pp.~15--26, ACM, 2017.

\bibitem{Madry2017}
A.~Madry, A.~Makelov, L.~Schmidt, D.~Tsipras, and A.~Vladu, ``Towards deep
  learning models resistant to adversarial attacks,'' in {\em International
  Conference on Learning Representations}, 2018.

\bibitem{dhillon2018stochastic}
G.~S. Dhillon, K.~Azizzadenesheli, Z.~C. Lipton, J.~Bernstein, J.~Kossaifi,
  A.~Khanna, and A.~Anandkumar, ``Stochastic activation pruning for robust
  adversarial defense,'' {\em arXiv preprint arXiv:1803.01442}, 2018.

\bibitem{metzen2017detecting}
J.~H. Metzen, T.~Genewein, V.~Fischer, and B.~Bischoff, ``On detecting
  adversarial perturbations,'' {\em arXiv preprint arXiv:1702.04267}, 2017.

\bibitem{grosse2017statistical}
K.~Grosse, P.~Manoharan, N.~Papernot, M.~Backes, and P.~McDaniel, ``On the
  (statistical) detection of adversarial examples,'' {\em arXiv preprint
  arXiv:1702.06280}, 2017.

\bibitem{galloway2017attacking}
A.~Galloway, G.~W. Taylor, and M.~Moussa, ``Attacking binarized neural
  networks,'' in {\em International Conference on Learning Representations},
  2018.

\bibitem{athalye2018obfuscated}
A.~Athalye, N.~Carlini, and D.~Wagner, ``Obfuscated gradients give a false
  sense of security: Circumventing defenses to adversarial examples,'' in {\em
  Proceedings of the 35th International Conference on Machine Learning, {ICML}
  2018}, 2018.

\bibitem{uesato2018adversarial}
J.~Uesato, B.~O'Donoghue, A.~v.~d. Oord, and P.~Kohli, ``Adversarial risk and
  the dangers of evaluating against weak attacks,'' in {\em Proceedings of the
  35th International Conference on Machine Learning, {ICML} 2018}, 2018.

\bibitem{lin2018defensive}
J.~Lin, C.~Gan, and S.~Han, ``Defensive quantization: When efficiency meets
  robustness,'' in {\em International Conference on Learning Representations},
  2019.

\bibitem{zhao2018compress}
Y.~Zhao, I.~Shumailov, R.~Mullins, and R.~Anderson, ``To compress or not to
  compress: Understanding the interactions between adversarial attacks and
  neural network compression,'' 2018.

\bibitem{rakin2018defend}
A.~S. Rakin, J.~Yi, B.~Gong, and D.~Fan, ``Defend deep neural networks against
  adversarial examples via fixed anddynamic quantized activation functions,''
  {\em arXiv preprint arXiv:1807.06714}, 2018.

\bibitem{Papernot2016}
N.~Papernot, P.~McDaniel, A.~Sinha, and M.~Wellman, ``Towards the science of
  security and privacy in machine learning,'' {\em arXiv preprint
  arXiv:1611.03814}, 2016.

\bibitem{khalil2018combinatorial}
E.~B. Khalil, A.~Gupta, and B.~Dilkina, ``Combinatorial attacks on binarized
  neural networks,'' {\em arXiv preprint arXiv:1810.03538}, 2018.

\bibitem{bengio2013estimating}
Y.~Bengio, N.~L{\'e}onard, and A.~Courville, ``Estimating or propagating
  gradients through stochastic neurons for conditional computation,'' {\em
  arXiv preprint arXiv:1308.3432}, 2013.

\bibitem{darabi2018bnn}
S.~Darabi, M.~Belbahri, M.~Courbariaux, and V.~P. Nia, ``Bnn+: Improved binary
  network training,'' 2018.

\bibitem{fredrikson2015model}
M.~Fredrikson, S.~Jha, and T.~Ristenpart, ``Model inversion attacks that
  exploit confidence information and basic countermeasures,'' in {\em
  Proceedings of the 22nd ACM SIGSAC Conference on Computer and Communications
  Security}, pp.~1322--1333, ACM, 2015.

\bibitem{shokri2017membership}
R.~Shokri, M.~Stronati, C.~Song, and V.~Shmatikov, ``Membership inference
  attacks against machine learning models,'' in {\em Security and Privacy (SP),
  2017 IEEE Symposium on}, pp.~3--18, IEEE, 2017.

\bibitem{shokri2015privacy}
R.~Shokri and V.~Shmatikov, ``Privacy-preserving deep learning,'' in {\em
  Proceedings of the 22nd ACM SIGSAC conference on computer and communications
  security}, pp.~1310--1321, ACM, 2015.

\bibitem{munoz2017towards}
L.~Mu{\~n}oz-Gonz{\'a}lez, B.~Biggio, A.~Demontis, A.~Paudice, V.~Wongrassamee,
  E.~C. Lupu, and F.~Roli, ``Towards poisoning of deep learning algorithms with
  back-gradient optimization,'' in {\em Proceedings of the 10th ACM Workshop on
  Artificial Intelligence and Security}, pp.~27--38, ACM, 2017.

\bibitem{yang2017generative}
C.~Yang, Q.~Wu, H.~Li, and Y.~Chen, ``Generative poisoning attack method
  against neural networks,'' {\em arXiv preprint arXiv:1703.01340}, 2017.

\bibitem{tanay2016boundary}
T.~Tanay and L.~Griffin, ``A boundary tilting persepective on the phenomenon of
  adversarial examples,'' {\em arXiv preprint arXiv:1608.07690}, 2016.

\bibitem{gilmer2018adversarial}
J.~Gilmer, L.~Metz, F.~Faghri, S.~S. Schoenholz, M.~Raghu, M.~Wattenberg, and
  I.~Goodfellow, ``Adversarial spheres,'' {\em arXiv preprint
  arXiv:1801.02774}, 2018.

\bibitem{samangouei2018defense}
P.~Samangouei, M.~Kabkab, and R.~Chellappa, ``Defense-gan: Protecting
  classifiers against adversarial attacks using generative models,'' {\em arXiv
  preprint arXiv:1805.06605}, 2018.

\bibitem{ilyas2019adversarial}
A.~Ilyas, S.~Santurkar, D.~Tsipras, L.~Engstrom, B.~Tran, and A.~Madry,
  ``Adversarial examples are not bugs, they are features,'' {\em arXiv preprint
  arXiv:1905.02175}, 2019.

\bibitem{kurakin2016adversarial2}
A.~Kurakin, I.~Goodfellow, and S.~Bengio, ``Adversarial examples in the
  physical world,'' in {\em International Conference on Learning
  Representations}, 2016.

\bibitem{spall1992multivariate}
J.~C. Spall {\em et~al.}, ``Multivariate stochastic approximation using a
  simultaneous perturbation gradient approximation,'' {\em IEEE transactions on
  automatic control}, vol.~37, no.~3, pp.~332--341, 1992.

\bibitem{serban2018adversarial}
A.~C. Serban and E.~Poll, ``Adversarial examples-a complete characterisation of
  the phenomenon,'' {\em arXiv preprint arXiv:1810.01185}, 2018.

\bibitem{buckman2018thermometer}
J.~Buckman, A.~Roy, C.~Raffel, and I.~Goodfellow, ``Thermometer encoding: One
  hot way to resist adversarial examples,'' in {\em International Conference on
  Learning Representations}, 2018.

\bibitem{xu2017feature}
W.~Xu, D.~Evans, and Y.~Qi, ``Feature squeezing: Detecting adversarial examples
  in deep neural networks,'' {\em arXiv preprint arXiv:1704.01155}, 2017.

\bibitem{xie2017mitigating}
C.~Xie, J.~Wang, Z.~Zhang, Z.~Ren, and A.~Yuille, ``Mitigating adversarial
  effects through randomization,'' {\em arXiv preprint arXiv:1711.01991}, 2017.

\bibitem{feinman2017detecting}
R.~Feinman, R.~R. Curtin, S.~Shintre, and A.~B. Gardner, ``Detecting
  adversarial samples from artifacts,'' {\em arXiv preprint arXiv:1703.00410},
  2017.

\bibitem{zheng2018robust}
Z.~Zheng and P.~Hong, ``Robust detection of adversarial attacks by modeling the
  intrinsic properties of deep neural networks,'' in {\em Advances in Neural
  Information Processing Systems}, pp.~7924--7933, 2018.

\bibitem{gong2017adversarial}
Z.~Gong, W.~Wang, and W.-S. Ku, ``Adversarial and clean data are not twins,''
  {\em arXiv preprint arXiv:1704.04960}, 2017.

\bibitem{meng2017magnet}
D.~Meng and H.~Chen, ``Magnet: a two-pronged defense against adversarial
  examples,'' in {\em Proceedings of the 2017 ACM SIGSAC Conference on Computer
  and Communications Security}, pp.~135--147, ACM, 2017.

\bibitem{lu2017safetynet}
J.~Lu, T.~Issaranon, and D.~A. Forsyth, ``Safetynet: Detecting and rejecting
  adversarial examples robustly.,'' in {\em ICCV}, pp.~446--454, 2017.

\bibitem{tramer2017ensemble}
F.~Tram{\`e}r, A.~Kurakin, N.~Papernot, I.~Goodfellow, D.~Boneh, and
  P.~McDaniel, ``Ensemble adversarial training: Attacks and defenses,'' {\em
  arXiv preprint arXiv:1705.07204}, 2017.

\bibitem{chang2018efficient}
T.-J. Chang, Y.~He, and P.~Li, ``Efficient two-step adversarial defense for
  deep neural networks,'' {\em arXiv preprint arXiv:1810.03739}, 2018.

\bibitem{kannan2018adversarial}
H.~Kannan, A.~Kurakin, and I.~Goodfellow, ``Adversarial logit pairing,'' {\em
  arXiv preprint arXiv:1803.06373}, 2018.

\bibitem{zheng2018pgdadversarial}
T.~Zheng, C.~Chen, and K.~Ren, ``Is pgd-adversarial training necessary?
  alternative training via a soft-quantization network with noisy-natural
  samples only,'' {\em arXiv preprint arXiv:1810.05665}, 2018.

\bibitem{zhang2019theoretically}
H.~Zhang, Y.~Yu, J.~Jiao, E.~P. Xing, L.~E. Ghaoui, and M.~I. Jordan,
  ``Theoretically principled trade-off between robustness and accuracy,'' {\em
  arXiv preprint arXiv:1901.08573}, 2019.

\bibitem{kariyappa2019improving}
S.~Kariyappa and M.~K. Qureshi, ``Improving adversarial robustness of ensembles
  with diversity training,'' {\em arXiv preprint arXiv:1901.09981}, 2019.

\bibitem{raghunathan2018certified}
A.~Raghunathan, J.~Steinhardt, and P.~Liang, ``Certified defenses against
  adversarial examples,'' {\em arXiv preprint arXiv:1801.09344}, 2018.

\bibitem{kolter2017provable}
J.~Z. Kolter and E.~Wong, ``Provable defenses against adversarial examples via
  the convex outer adversarial polytope,'' {\em arXiv preprint
  arXiv:1711.00851}, vol.~1, no.~2, p.~3, 2017.

\bibitem{hein2017formal}
M.~Hein and M.~Andriushchenko, ``Formal guarantees on the robustness of a
  classifier against adversarial manipulation,'' in {\em Advances in Neural
  Information Processing Systems}, pp.~2266--2276, 2017.

\bibitem{peck2017lower}
J.~Peck, J.~Roels, B.~Goossens, and Y.~Saeys, ``Lower bounds on the robustness
  to adversarial perturbations,'' in {\em Advances in Neural Information
  Processing Systems}, pp.~804--813, 2017.

\bibitem{gowal2018effectiveness}
S.~Gowal, K.~Dvijotham, R.~Stanforth, R.~Bunel, C.~Qin, J.~Uesato, T.~Mann, and
  P.~Kohli, ``On the effectiveness of interval bound propagation for training
  verifiably robust models,'' {\em arXiv preprint arXiv:1810.12715}, 2018.

\bibitem{kingma2014adam}
D.~P. Kingma and J.~Ba, ``Adam: A method for stochastic optimization,'' {\em
  arXiv preprint arXiv:1412.6980}, 2014.

\bibitem{papernot2018cleverhans}
N.~Papernot, F.~Faghri, N.~Carlini, I.~Goodfellow, R.~Feinman, A.~Kurakin,
  C.~Xie, Y.~Sharma, T.~Brown, A.~Roy, A.~Matyasko, V.~Behzadan,
  K.~Hambardzumyan, Z.~Zhang, Y.-L. Juang, Z.~Li, R.~Sheatsley, A.~Garg,
  J.~Uesato, W.~Gierke, Y.~Dong, D.~Berthelot, P.~Hendricks, J.~Rauber, and
  R.~Long, ``Technical report on the cleverhans v2.1.0 adversarial examples
  library,'' {\em arXiv preprint arXiv:1610.00768}, 2018.

\bibitem{carlini2019evaluating}
N.~Carlini, A.~Athalye, N.~Papernot, W.~Brendel, J.~Rauber, D.~Tsipras,
  I.~Goodfellow, and A.~Madry, ``On evaluating adversarial robustness,'' {\em
  arXiv preprint arXiv:1902.06705}, 2019.

\bibitem{wu2018understanding}
L.~Wu, Z.~Zhu, C.~Tai, {\em et~al.}, ``Understanding and enhancing the
  transferability of adversarial examples,'' {\em arXiv preprint
  arXiv:1802.09707}, 2018.

\bibitem{ilyas2017robust}
A.~Ilyas, A.~Jalal, E.~Asteri, C.~Daskalakis, and A.~G. Dimakis, ``The robust
  manifold defense: Adversarial training using generative models,'' {\em arXiv
  preprint arXiv:1712.09196}, 2017.

\bibitem{carlini2017adversarial}
N.~Carlini and D.~Wagner, ``Adversarial examples are not easily detected:
  Bypassing ten detection methods,'' in {\em Proceedings of the 10th ACM
  Workshop on Artificial Intelligence and Security}, pp.~3--14, ACM, 2017.

\end{thebibliography}

\newpage

\appendix
\section{Networks architecture}
\label{Network architecture}

\begin{table}[h!]
\centering
\begin{tabularx}{1.1\linewidth}{lllllll}
    \hline
    \multicolumn{2}{l}{Layer type} & \multicolumn{2}{l}{CIFAR10} & \multicolumn{2}{l}{SVHN}\\
    Convolution + BatchNorm + relu & & (128,3,3) & & (128,3,3) &\\
    Convolution + MaxPooling + BatchNorm + relu & & (128,3,3), (2,2) & & (128,3,3), (2,2) &\\
    Convolution + BatchNorm + relu & & (256,3,3) & & (256,3,3) &\\
    Convolution + MaxPooling + BatchNorm + relu & & (256,3,3), (2,2) & & (128,3,3), (2,2) &\\
    Convolution + BatchNorm + relu & & (512,3,3) & & (512,3,3) &\\
    Convolution + MaxPooling + BatchNorm + relu \quad\quad\quad\quad\quad\quad\quad\quad& & (512,3,3), (2,2) & & (512,3,3), (2,2) &\\
    Fully Connected + BatchNorm + relu & & 1024, (2,2) & & 1024, (2,2) &\\
    Fully Connected + BatchNorm + relu & & 1024, (2,2) & & 1024, (2,2) &\\
    Fully Connected + softmax & & 10 & & 10 &\\
    \hline
\end{tabularx}
\caption{Full-precision models architecture}
\end{table}

\begin{table}[h!]
\centering
\begin{tabularx}{1.1\linewidth}{lllllll}
    \hline
    \multicolumn{2}{l}{Layer type} & \multicolumn{2}{l}{CIFAR10} & \multicolumn{2}{l}{SVHN}\\
    ConvolutionQuant + BatchNorm + reluQuant & & (128,3,3) & & (128,3,3) &\\
    ConvolutionreluQuant + MaxPooling + BatchNorm + reluQuant & & (128,3,3), (2,2) & & (128,3,3), (2,2) &\\
    ConvolutionQuant + BatchNorm + reluQuant & & (256,3,3) & & (256,3,3) &\\
    ConvolutionQuant + MaxPooling + BatchNorm + reluQuant & & (256,3,3), (2,2) & & (128,3,3), (2,2) &\\
    ConvolutionQuant + BatchNorm + reluQuant & & (512,3,3) & & (512,3,3) &\\
    ConvolutionQuant + MaxPooling + BatchNorm + reluQuant & & (512,3,3), (2,2) & & (512,3,3), (2,2) &\\
    DenseQuant + BatchNorm + reluQuant & & 1024, (2,2) & & 1024, (2,2) &\\
    DenseQuant + BatchNorm + reluQuant & & 1024, (2,2) & & 1024, (2,2) &\\
    Dense + softmax & & 10 & & 10 &\\
    \hline
\end{tabularx}
\caption{Fully Quantized models architecture. ConvolutionQuant, DenseQuant and reluQuant designate respectively a convolution layer with quantized weights, a dense layer with quantized weights and the relu activation function with its output quantized}
\end{table}

\begin{table}[h!]
\centering
\begin{tabularx}{1.1\linewidth}{lllllll}
    \hline
    \multicolumn{2}{l}{Layer type} & \multicolumn{2}{l}{CIFAR10} & \multicolumn{2}{l}{SVHN}\\
    ConvolutionQuant + BatchNorm + reluQuant & & (128,3,3) & & (128,3,3) &\\
    ConvolutionreluQuant + MaxPooling + BatchNorm + relu & & (128,3,3), (2,2) & & (128,3,3), (2,2) &\\
    ConvolutionQuant + BatchNorm + reluQuant & & (256,3,3) & & (256,3,3) &\\
    ConvolutionQuant + MaxPooling + BatchNorm + relu & & (256,3,3), (2,2) & & (128,3,3), (2,2) &\\
    ConvolutionQuant + BatchNorm + reluQuant & & (512,3,3) & & (512,3,3) &\\
    ConvolutionQuant + MaxPooling + BatchNorm + relu \quad\quad\quad\quad\quad& & (512,3,3), (2,2) & & (512,3,3), (2,2) &\\
    DenseQuant + BatchNorm + relu & & 1024, (2,2) & & 1024, (2,2) &\\
    DenseQuant + BatchNorm + relu & & 1024, (2,2) & & 1024, (2,2) &\\
    Dense + softmax & & 10 & & 10 &\\
    \hline
\end{tabularx}
\caption{Weight Quantized models architecture. ConvolutionQuant and DenseQuant designate respectively a convolution layer with quantized weights and a dense layer with quantized weights}
\end{table}

\newpage
\section{Attacks parameters}
\label{Attack parameters}

For ZOO and CWl2, we noticed that results between 100 and 1000 iterations were almost similar, the adversarial accuracy almost never decreased and the $l_2$ distortion for the two attacks decreased proportionally. For computation time issues we then chose to perform the attack with 100 iterations, as this does not change any interpretations of our results.\\

The value of $\kappa$ for the CWl2 attack is set to $0$ when considering an adversary in the white-box setting (see Section \ref{Robustness against gradient-based and gradient-free attacks}). Otherwise, in particular for transfer-based attacks, this parameter is tuned (see Section \ref{Transferability} for details).

\begin{table}[h!]
\centering
\begin{tabularx}{0.5\linewidth}{XX}
    \hline
    $\epsilon$ & 0.03\\
    \hline
\end{tabularx}
\caption{Hyperparameters for FGSM}
\end{table}
      
\begin{table}[h!]
\centering
\begin{tabularx}{0.5\linewidth}{XX}
    \hline
    $\epsilon$ & 0.03\\
    \textbf{Iterations} & 100 \\
    \textbf{Step size} & 0.0003 \\
    \hline
\end{tabularx}
\caption{Hyperparameters for BIM}
\end{table}

\begin{table}[h!]
\centering
\begin{tabularx}{0.5\linewidth}{XX}
    \hline
    $\epsilon$ & 0.03\\
    \textbf{Iterations} & 100\\
    \textbf{Learning rate} & 0.01\\
    \textbf{Perturbation size} $\delta$ & 0.01\\
    \textbf{Batch size} & 128 \\
    \hline
\end{tabularx}
\caption{Hyperparameters for SPSA}
\end{table}

\begin{table}[h!]
\centering
\begin{tabularx}{0.5\linewidth}{XX}
    \hline
    \textbf{Iterations} & 100\\
    \textbf{Learning rate} & 0.1\\
    \textbf{Initial constant} & 0.9\\
    \textbf{Search steps} & 10 \\
    $\boldsymbol{\kappa}$ & 0 \\
    \hline
\end{tabularx}
\caption{Hyperparameters for CWl2}
\end{table}

\begin{table}[h!]
\centering
\begin{tabularx}{0.5\linewidth}{XX}
    \hline
    \textbf{Iterations} & 100\\
    \textbf{Learning rate} & 0.1\\
    \textbf{Initial constant} & 0.9\\
    \textbf{Search steps} & 10 \\
    $\boldsymbol{\kappa}$ & 0 \\
    \hline
\end{tabularx}
\caption{Hyperparameters for ZOO}
\end{table}

\newpage
\section{Complete transferability results}     
\label{Black-box other results}   

In the following tables, "--" denotes a value which can not be computed. For example, the $l_2$ distortion of successful adversarial examples for an attack can not be computed when the adversarial accuracy of the target model against this attack equals 1.

We summarize the reference of the transferability tables where $w_ia_j$ designates a model with a \textit{i}-bit quantization of the weights and a \textit{j}-bit quantization of the activation values.

\begin{table}[h!]
\begin{center}
\scalebox{0.75}{
\begin{tabular}{|c|ccccc|cccc|}
 \hline
 
 \diagbox{From}{To} & full & $w_1a_1$ & $w_2a_2$ & $w_3a_3$ & $w_4a_4$ & $w_1a_{32}$ & $w_2a_{32}$ & $w_3a_{32}$ & $w_4a_{32}$
 \\
 \hline
 
 full&\multicolumn{5}{c|}{Table~\ref{Float to fully}} & \multicolumn{4}{c|}{Table~\ref{Float to weight}}
\\
 $w_1a_1$&\multicolumn{5}{c|}{Table~\ref{w1 a1 to fully}} & \multicolumn{4}{c|}{Table~\ref{w1 a1 to weight}}
\\
 $w_1a_{32}$&\multicolumn{5}{c|}{Table~\ref{w1 a32 to fully}} & \multicolumn{4}{c|}{Table~\ref{w1 a32 to weight}}
\\
 $w_2a_2$ &\multicolumn{5}{c|}{Table~\ref{w2 a2 to fully}} & \multicolumn{4}{c|}{Table~\ref{w2 a2 to weight}}
\\
 $w_2a_{32}$&\multicolumn{5}{c|}{Table~\ref{w2 a32 to fully}} & \multicolumn{4}{c|}{Table~\ref{w2 a32 to weight}}
\\
 $w_3a_3$&\multicolumn{5}{c|}{Table~\ref{w3 a3 to fully}} & \multicolumn{4}{c|}{Table~\ref{w3 a3 to weight}}
\\
 $w_3a_{32}$&\multicolumn{5}{c|}{Table~\ref{w3 a32 to fully}} & \multicolumn{4}{c|}{Table~\ref{w3 a32 to weight}}
\\
 $w_4a_4$&\multicolumn{5}{c|}{Table~\ref{w4 a4 to fully}} & \multicolumn{4}{c|}{Table~\ref{w4 a4 to weight}}
\\
 $w_4a_{32}$&\multicolumn{5}{c|}{Table~\ref{w4 a32 to fully}} & \multicolumn{4}{c|}{Table~\ref{w4 a32 to weight}}\\
 \hline 
 \end{tabular}
 }
\end{center}
\caption{Summary of the references for the transferability results between full precision, fully quantized and weight only models.}
\label{ref_trans_tables}
\end{table}                          
               
\begin{table}[h!]
\begin{center}
\scalebox{0.9}{
\begin{tabularx}{\linewidth}{ccccccccccccc}
 \hline
 \\~\vspace{-20pt}\\
 \multirow{4}{*}{~} & \multicolumn{6}{c}{CIFAR10} & \multicolumn{6}{c}{SVHN}

 \\~\vspace{-20pt}\\
 \\&\multicolumn{3}{c}{Float model} & \multicolumn{3}{c}{Quantized models} & \multicolumn{3}{c}{Float model} & \multicolumn{3}{c}{Quantized models}

 \\&\multicolumn{3}{c}{\textit{(32-bit)}} & \multicolumn{3}{c}{\textit{(1,2,3,4-bit)}} & \multicolumn{3}{c}{\textit{(32-bit)}} & \multicolumn{3}{c}{\textit{(1,2,3,4-bit)}}

 \\~\vspace{-20pt}\\
\hline 
 \end{tabularx}
 }
\end{center}
\caption{Transferability from full-precision model to 1,2,3,4-bit fully quantized models.}
\label{Float to fully}
\end{table}


 \begin{table}[h!]
\begin{center}
\scalebox{0.9}{
\begin{tabularx}{\linewidth}{ccccccccccccc}
 \hline
 \\~\vspace{-20pt}\\
 \multirow{4}{*}{~} & \multicolumn{6}{c}{CIFAR10} & \multicolumn{6}{c}{SVHN}

 \\~\vspace{-20pt}\\
 \\&\multicolumn{3}{c}{Float model} & \multicolumn{3}{c}{Quantized models} & \multicolumn{3}{c}{Float model} & \multicolumn{3}{c}{Quantized models}

 \\&\multicolumn{3}{c}{\textit{(32-bit)}} & \multicolumn{3}{c}{\textit{(1,2,3,4-bit)}} & \multicolumn{3}{c}{\textit{(32-bit)}} & \multicolumn{3}{c}{\textit{(1,2,3,4-bit)}}

 \\~\vspace{-20pt}\\
\hline 
 \end{tabularx}
}
\end{center}
\caption{Transferability from full-precision model to 1,2,3,4-bit weight-only quantized models.}
\label{Float to weight}
\end{table}

 \begin{table}[h!]
\begin{center}
\scalebox{0.9}{
\begin{tabularx}{\linewidth}{ccccccccccccc}
 \hline
 \\~\vspace{-20pt}\\
 \multirow{4}{*}{~} & \multicolumn{6}{c}{CIFAR10} & \multicolumn{6}{c}{SVHN}

 \\~\vspace{-20pt}\\
 \\&\multicolumn{3}{c}{Float model} & \multicolumn{3}{c}{Quantized models} & \multicolumn{3}{c}{Float model} & \multicolumn{3}{c}{Quantized models}

 \\&\multicolumn{3}{c}{\textit{(32-bit)}} & \multicolumn{3}{c}{\textit{(1,2,3,4-bit)}} & \multicolumn{3}{c}{\textit{(32-bit)}} & \multicolumn{3}{c}{\textit{(1,2,3,4-bit)}}

 \\~\vspace{-20pt}\\
\hline 
 \end{tabularx}
 }
\end{center}
\caption{Transferability from fully binarized model to 1,2,3,4-bit fully quantized models and full-precision models.}
\label{w1 a1 to fully}
\end{table}

\begin{table}[h!]
\begin{center}
\scalebox{0.9}{
\begin{tabularx}{\linewidth}{ccccccccccccc}
 \hline
 \\~\vspace{-20pt}\\
 \multirow{4}{*}{~} & \multicolumn{6}{c}{CIFAR10} & \multicolumn{6}{c}{SVHN}

 \\~\vspace{-20pt}\\
 \\&\multicolumn{3}{c}{Float model} & \multicolumn{3}{c}{Quantized models} & \multicolumn{3}{c}{Float model} & \multicolumn{3}{c}{Quantized models}

 \\&\multicolumn{3}{c}{\textit{(32-bit)}} & \multicolumn{3}{c}{\textit{(1,2,3,4-bit)}} & \multicolumn{3}{c}{\textit{(32-bit)}} & \multicolumn{3}{c}{\textit{(1,2,3,4-bit)}}

 \\~\vspace{-20pt}\\
\hline
 \end{tabularx}
 }
\end{center}
\caption{Transferability from fully binarized model to 1,2,3,4-bit weight-only quantized models and full-precision models.}
\label{w1 a1 to weight}
\end{table}

\begin{table}[h!]
\begin{center}
\scalebox{0.9}{
\begin{tabularx}{\linewidth}{ccccccccccccc}
 \hline
 \\~\vspace{-20pt}\\
 \multirow{4}{*}{~} & \multicolumn{6}{c}{CIFAR10} & \multicolumn{6}{c}{SVHN}

 \\~\vspace{-20pt}\\
 \\&\multicolumn{3}{c}{Float model} & \multicolumn{3}{c}{Quantized models} & \multicolumn{3}{c}{Float model} & \multicolumn{3}{c}{Quantized models}

 \\&\multicolumn{3}{c}{\textit{(32-bit)}} & \multicolumn{3}{c}{\textit{(1,2,3,4-bit)}} & \multicolumn{3}{c}{\textit{(32-bit)}} & \multicolumn{3}{c}{\textit{(1,2,3,4-bit)}}

 \\~\vspace{-20pt}\\
\hline
\end{tabularx}
}
\end{center}
\caption{Transferability from weight-only binarized model to 1,2,3,4-bit fully quantized models and full-precision models.}
\label{w1 a32 to fully}
\end{table}

 \begin{table}[h!]
\begin{center}
\scalebox{0.9}{
\begin{tabularx}{\linewidth}{ccccccccccccc}
 \hline
 \\~\vspace{-20pt}\\
 \multirow{4}{*}{~} & \multicolumn{6}{c}{CIFAR10} & \multicolumn{6}{c}{SVHN}

 \\~\vspace{-20pt}\\
 \\&\multicolumn{3}{c}{Float model} & \multicolumn{3}{c}{Quantized models} & \multicolumn{3}{c}{Float model} & \multicolumn{3}{c}{Quantized models}

 \\&\multicolumn{3}{c}{\textit{(32-bit)}} & \multicolumn{3}{c}{\textit{(1,2,3,4-bit)}} & \multicolumn{3}{c}{\textit{(32-bit)}} & \multicolumn{3}{c}{\textit{(1,2,3,4-bit)}}

 \\~\vspace{-20pt}\\
\hline 
 \end{tabularx}
 }
\end{center}
\caption{Transferability from  weight-only binarized model to 1,2,3,4-bit weight-only quantized models and full-precision models.}
\label{w1 a32 to weight}
\end{table}

\begin{table}[h!]
\begin{center}
\scalebox{0.9}{
\begin{tabularx}{\linewidth}{ccccccccccccc}
 \hline
 \\~\vspace{-20pt}\\
 \multirow{4}{*}{~} & \multicolumn{6}{c}{CIFAR10} & \multicolumn{6}{c}{SVHN}

 \\~\vspace{-20pt}\\
 \\&\multicolumn{3}{c}{Float model} & \multicolumn{3}{c}{Quantized models} & \multicolumn{3}{c}{Float model} & \multicolumn{3}{c}{Quantized models}

 \\&\multicolumn{3}{c}{\textit{(32-bit)}} & \multicolumn{3}{c}{\textit{(1,2,3,4-bit)}} & \multicolumn{3}{c}{\textit{(32-bit)}} & \multicolumn{3}{c}{\textit{(1,2,3,4-bit)}}

 \\~\vspace{-20pt}\\
\hline 
 \end{tabularx}
 }
\end{center}
\caption{Transferability from  2-bit fully quantized model to 1,2,3,4-bit fully quantized models.}
\label{w2 a2 to fully}
\end{table}

 \begin{table}[h!]
\begin{center}
\scalebox{0.9}{
\begin{tabularx}{\linewidth}{ccccccccccccc}
 \hline
 \\~\vspace{-20pt}\\
 \multirow{4}{*}{~} & \multicolumn{6}{c}{CIFAR10} & \multicolumn{6}{c}{SVHN}

 \\~\vspace{-20pt}\\
 \\&\multicolumn{3}{c}{Float model} & \multicolumn{3}{c}{Quantized models} & \multicolumn{3}{c}{Float model} & \multicolumn{3}{c}{Quantized models}

 \\&\multicolumn{3}{c}{\textit{(32-bit)}} & \multicolumn{3}{c}{\textit{(1,2,3,4-bit)}} & \multicolumn{3}{c}{\textit{(32-bit)}} & \multicolumn{3}{c}{\textit{(1,2,3,4-bit)}}

 \\~\vspace{-20pt}\\
\hline 
 \end{tabularx}
 }
\end{center}
\caption{Transferability from  2-bit fully quantized model to 1,2,3,4-bit weight-only quantized models.}
\label{w2 a2 to weight}
\end{table}

\begin{table}[h!]
\begin{center}
\scalebox{0.9}{
\begin{tabularx}{\linewidth}{ccccccccccccc}
 \hline
 \\~\vspace{-20pt}\\
 \multirow{4}{*}{~} & \multicolumn{6}{c}{CIFAR10} & \multicolumn{6}{c}{SVHN}

 \\~\vspace{-20pt}\\
 \\&\multicolumn{3}{c}{Float model} & \multicolumn{3}{c}{Quantized models} & \multicolumn{3}{c}{Float model} & \multicolumn{3}{c}{Quantized models}

 \\&\multicolumn{3}{c}{\textit{(32-bit)}} & \multicolumn{3}{c}{\textit{(1,2,3,4-bit)}} & \multicolumn{3}{c}{\textit{(32-bit)}} & \multicolumn{3}{c}{\textit{(1,2,3,4-bit)}}
 
 \\~\vspace{-20pt}\\
\hline 
 \end{tabularx}
 }
\end{center}
\caption{Transferability from  2-bit weight-only quantized model to 1,2,3,4-bit fully quantized models.}
\label{w2 a32 to fully}
\end{table} 

\begin{table}[h!]
\begin{center}
\scalebox{0.9}{
\begin{tabularx}{\linewidth}{ccccccccccccc}
 \hline
 \\~\vspace{-20pt}\\
 \multirow{4}{*}{~} & \multicolumn{6}{c}{CIFAR10} & \multicolumn{6}{c}{SVHN}

 \\~\vspace{-20pt}\\
 \\&\multicolumn{3}{c}{Float model} & \multicolumn{3}{c}{Quantized models} & \multicolumn{3}{c}{Float model} & \multicolumn{3}{c}{Quantized models}

 \\&\multicolumn{3}{c}{\textit{(32-bit)}} & \multicolumn{3}{c}{\textit{(1,2,3,4-bit)}} & \multicolumn{3}{c}{\textit{(32-bit)}} & \multicolumn{3}{c}{\textit{(1,2,3,4-bit)}}
 
 \\~\vspace{-20pt}\\
\hline 
 \end{tabularx}
 }
\end{center}
\caption{Transferability from  2-bit weight-only quantized model to 1,2,3,4-bit weight-only quantized models.}
\label{w2 a32 to weight}
\end{table}                
               
\begin{table}[h!]
\begin{center}
\scalebox{0.9}{
\begin{tabularx}{\linewidth}{ccccccccccccc}
 \hline
 \\~\vspace{-20pt}\\
 \multirow{4}{*}{~} & \multicolumn{6}{c}{CIFAR10} & \multicolumn{6}{c}{SVHN}

 \\~\vspace{-20pt}\\
 \\&\multicolumn{3}{c}{Float model} & \multicolumn{3}{c}{Quantized models} & \multicolumn{3}{c}{Float model} & \multicolumn{3}{c}{Quantized models}

 \\&\multicolumn{3}{c}{\textit{(32-bit)}} & \multicolumn{3}{c}{\textit{(1,2,3,4-bit)}} & \multicolumn{3}{c}{\textit{(32-bit)}} & \multicolumn{3}{c}{\textit{(1,2,3,4-bit)}}

 \\~\vspace{-20pt}\\
\hline 
 \end{tabularx}
 }
\end{center}
\caption{Transferability from 3-bit fully quantized model to 1,2,3,4-bit fully quantized models.}
\label{w3 a3 to fully}
\end{table}

 \begin{table}[h!]
\begin{center}
\scalebox{0.9}{
\begin{tabularx}{\linewidth}{ccccccccccccc}
 \hline
 \\~\vspace{-20pt}\\
 \multirow{4}{*}{~} & \multicolumn{6}{c}{CIFAR10} & \multicolumn{6}{c}{SVHN}

 \\~\vspace{-20pt}\\
 \\&\multicolumn{3}{c}{Float model} & \multicolumn{3}{c}{Quantized models} & \multicolumn{3}{c}{Float model} & \multicolumn{3}{c}{Quantized models}

 \\&\multicolumn{3}{c}{\textit{(32-bit)}} & \multicolumn{3}{c}{\textit{(1,2,3,4-bit)}} & \multicolumn{3}{c}{\textit{(32-bit)}} & \multicolumn{3}{c}{\textit{(1,2,3,4-bit)}}

 \\~\vspace{-20pt}\\
\hline 
 \end{tabularx}
 }
\end{center}
\caption{Transferability from  3-bit fully quantized model to 1,2,3,4-bit weight-only quantized models.}
\label{w3 a3 to weight}
\end{table}

 \begin{table}[h!]
\begin{center}
\scalebox{0.9}{
\begin{tabularx}{\linewidth}{ccccccccccccc}
 \hline
 \\~\vspace{-20pt}\\
 \multirow{4}{*}{~} & \multicolumn{6}{c}{CIFAR10} & \multicolumn{6}{c}{SVHN}

 \\~\vspace{-20pt}\\
 \\&\multicolumn{3}{c}{Float model} & \multicolumn{3}{c}{Quantized models} & \multicolumn{3}{c}{Float model} & \multicolumn{3}{c}{Quantized models}

 \\&\multicolumn{3}{c}{\textit{(32-bit)}} & \multicolumn{3}{c}{\textit{(1,2,3,4-bit)}} & \multicolumn{3}{c}{\textit{(32-bit)}} & \multicolumn{3}{c}{\textit{(1,2,3,4-bit)}}

 \\~\vspace{-20pt}\\
\hline 
\end{tabularx}
}
\end{center}
\caption{Transferability from 3-bit weight-only quantized model to 1,2,3,4-bit fully quantized models.}
\label{w3 a32 to fully}
\end{table}

\begin{table}[h!]
\begin{center}
\scalebox{0.9}{
\begin{tabularx}{\linewidth}{ccccccccccccc}
 \hline
 \\~\vspace{-20pt}\\
 \multirow{4}{*}{~} & \multicolumn{6}{c}{CIFAR10} & \multicolumn{6}{c}{SVHN}

 \\~\vspace{-20pt}\\
 \\&\multicolumn{3}{c}{Float model} & \multicolumn{3}{c}{Quantized models} & \multicolumn{3}{c}{Float model} & \multicolumn{3}{c}{Quantized models}

 \\&\multicolumn{3}{c}{\textit{(32-bit)}} & \multicolumn{3}{c}{\textit{(1,2,3,4-bit)}} & \multicolumn{3}{c}{\textit{(32-bit)}} & \multicolumn{3}{c}{\textit{(1,2,3,4-bit)}}

 \\~\vspace{-20pt}\\
\hline
 \end{tabularx}
 }
\end{center}
\caption{Transferability from  3-bit weight-only quantized model to 1,2,3,4-bit weight-only quantized models.}
\label{w3 a32 to weight}
\end{table}

\begin{table}[h!]
\begin{center}
\scalebox{0.9}{
\begin{tabularx}{\linewidth}{ccccccccccccc}
 \hline
 \\~\vspace{-20pt}\\
 \multirow{4}{*}{~} & \multicolumn{6}{c}{CIFAR10} & \multicolumn{6}{c}{SVHN}

 \\~\vspace{-20pt}\\
 \\&\multicolumn{3}{c}{Float model} & \multicolumn{3}{c}{Quantized models} & \multicolumn{3}{c}{Float model} & \multicolumn{3}{c}{Quantized models}

 \\&\multicolumn{3}{c}{\textit{(32-bit)}} & \multicolumn{3}{c}{\textit{(1,2,3,4-bit)}} & \multicolumn{3}{c}{\textit{(32-bit)}} & \multicolumn{3}{c}{\textit{(1,2,3,4-bit)}}
 
 \\~\vspace{-20pt}\\
\hline 
 \end{tabularx}
 }
\end{center}
\caption{Transferability from  4-bit fully quantized model to 1,2,3,4-bit fully quantized models.}
\label{w4 a4 to fully}
\end{table} 

\begin{table}[h!]
\begin{center}
\scalebox{0.9}{
\begin{tabularx}{\linewidth}{ccccccccccccc}
 \hline
 \\~\vspace{-20pt}\\
 \multirow{4}{*}{~} & \multicolumn{6}{c}{CIFAR10} & \multicolumn{6}{c}{SVHN}

 \\~\vspace{-20pt}\\
 \\&\multicolumn{3}{c}{Float model} & \multicolumn{3}{c}{Quantized models} & \multicolumn{3}{c}{Float model} & \multicolumn{3}{c}{Quantized models}

 \\&\multicolumn{3}{c}{\textit{(32-bit)}} & \multicolumn{3}{c}{\textit{(1,2,3,4-bit)}} & \multicolumn{3}{c}{\textit{(32-bit)}} & \multicolumn{3}{c}{\textit{(1,2,3,4-bit)}}
 
 \\~\vspace{-20pt}\\
\hline 
 \end{tabularx}
 }
\end{center}
\caption{Transferability from  4-bit fully quantized model to 1,2,3,4-bit weight-only quantized models.}
\label{w4 a4 to weight}
\end{table} 

\begin{table}[h!]
\begin{center}
\scalebox{0.9}{
\begin{tabularx}{\linewidth}{ccccccccccccc}
 \hline
 \\~\vspace{-20pt}\\
 \multirow{4}{*}{~} & \multicolumn{6}{c}{CIFAR10} & \multicolumn{6}{c}{SVHN}

 \\~\vspace{-20pt}\\
 \\&\multicolumn{3}{c}{Float model} & \multicolumn{3}{c}{Quantized models} & \multicolumn{3}{c}{Float model} & \multicolumn{3}{c}{Quantized models}

 \\&\multicolumn{3}{c}{\textit{(32-bit)}} & \multicolumn{3}{c}{\textit{(1,2,3,4-bit)}} & \multicolumn{3}{c}{\textit{(32-bit)}} & \multicolumn{3}{c}{\textit{(1,2,3,4-bit)}}
 
 \\~\vspace{-20pt}\\
\hline 
 \end{tabularx}
 }
\end{center}
\caption{Transferability from  4-bit weight-only quantized model to 1,2,3,4-bit fully quantized models.}
\label{w4 a32 to fully}
\end{table} 

\begin{table}[h!]
\begin{center}
\scalebox{0.9}{
\begin{tabularx}{\linewidth}{ccccccccccccc}
 \hline
 \\~\vspace{-20pt}\\
 \multirow{4}{*}{~} & \multicolumn{6}{c}{CIFAR10} & \multicolumn{6}{c}{SVHN}

 \\~\vspace{-20pt}\\
 \\&\multicolumn{3}{c}{Float model} & \multicolumn{3}{c}{Quantized models} & \multicolumn{3}{c}{Float model} & \multicolumn{3}{c}{Quantized models}

 \\&\multicolumn{3}{c}{\textit{(32-bit)}} & \multicolumn{3}{c}{\textit{(1,2,3,4-bit)}} & \multicolumn{3}{c}{\textit{(32-bit)}} & \multicolumn{3}{c}{\textit{(1,2,3,4-bit)}}
 
 \\~\vspace{-20pt}\\
\hline 
 \end{tabularx}
 }
\end{center}
\caption{Transferability from  4-bit weight-only quantized model to 1,2,3,4-bit weight-only quantized models.}
\label{w4 a32 to weight}
\end{table}

\end{document}